\documentclass{article}

\pdfoutput=1

\usepackage{arxiv}
\usepackage[utf8]{inputenc} % allow utf-8 input
\usepackage[T1]{fontenc}    % use 8-bit T1 fonts
\usepackage{hyperref}       % hyperlinks
\usepackage{url}            % simple URL typesetting
\usepackage{booktabs}       % professional-quality tables
\usepackage{amsfonts}       % blackboard math symbols
\usepackage{nicefrac}       % compact symbols for 1/2, etc.
\usepackage{microtype}      % microtypography
\usepackage{multirow}
\usepackage{multicol}
\usepackage{subfigure}
\usepackage{caption}
\usepackage{graphicx}
\usepackage{mathrsfs}
\usepackage{algorithm}
\usepackage{algorithmic}
\usepackage{bm}
\usepackage{color}
\usepackage{authblk}
\title{Learning to Purify Noisy Labels via Meta Soft Label Corrector}
\date{}
\author[]{Yichen Wu}
\author[]{Jun Shu}
\author[]{Qi Xie}
\author[]{Qian Zhao}
\author[]{Deyu Meng\thanks{Corresponding Author}}
\affil[]{School of Mathematics and Statistics, Xi'an Jiaotong University}
\begin{document}

\maketitle
\vspace{-7mm}
\begin{abstract}
Recent deep neural networks (DNNs) can easily overfit to biased training data with noisy labels. Label correction strategy is commonly used to alleviate this issue by designing a method to identity suspected noisy labels and then correct them. Current approaches to correcting corrupted labels usually need certain pre-defined label correction rules or manually preset hyper-parameters. These fixed settings make it hard to apply in practice since the accurate label correction usually related with the concrete problem, training data and the temporal information hidden in dynamic iterations of training process. To address this issue, we propose a meta-learning model which could estimate soft labels through meta-gradient descent step under the guidance of noise-free meta data. By viewing the label correction procedure as a meta-process and using a meta-learner to automatically correct labels, we could adaptively obtain rectified soft labels iteratively according to current training problems without manually preset hyper-parameters. Besides, our method is model-agnostic and we can combine it with any other existing model with ease. Comprehensive experiments substantiate the superiority of our method in both synthetic and real-world problems with noisy labels compared with current SOTA label correction strategies.
\end{abstract}
\vspace{-4.2mm}
\section{Introduction}
The remarkable success of deep neural networks (DNNs) on various tasks heavily relies on pre-collected large-scale dataset with high-quality annotations \cite{he2016deep,krizhevsky2012imagenet}. However, practical annotated training dataset always contains certain amount of noisy (incorrect) labels, easily conducting overfitting issue and leading to the poor performance of the trained DNNs in generalization \cite{zhang2016understanding}. In fact, such biased training data are commonly encountered in practice, due to the coarse annotation sources for collecting them, like web searches \cite{liu2011noise} and crowd-sourcing \cite{welinder2010multidimensional}. Therefore, how to train DNNs robustly with such biased training data is a critical issue in current machine learning field.
%In order to tackle this challenge

To address this problem, various methods have been proposed \cite{arazo2019unsupervised,shu2019meta,jiang2018mentornet}, which can be coarsely categorized as sample selection and label correction approaches.
Sample selection approaches tackle this challenge mainly via adopting sample re-weighting schemes by imposing importance weights on samples according to their loss. Typical methods include boosting and self-paced learning methods \cite{kumar2010self,meng2017theoretical}. Recently, some pioneering works \cite{ren2018learning,shu2019meta} further make such weighting schemes more adaptive and automatic through employing a small set of clean meta data to guide the network training process. All these methods are built on the basis of throwing off the suspected noisy samples in training. However, these corrupted samples contain beneficial information that could improve the accuracy and robustness of network especially in large noise-ratio scenarios \cite{chang2017active}.

Label correction approaches alleviate this issue through attempting to find and correct noisy labels to their underlying correct ones. For example, \cite{hendrycks2018using}\cite{patrini2017making}\cite{shu2020meta} revised class probabilities through estimating noise transition matrix, aiming to recover the underlying ground-truth label distribution to guide the training process towards the correct classes. However, owing to the difficulty in estimating the noise transition matrix or true labels, the network training could easily accumulate errors, especially when the number of classes or mislabeled samples is large \cite{shu2019meta,jiang2018mentornet}. Another common methodology is to directly rectify the noisy labels by exploiting the prediction of network, e.g., Reed et al.\cite{reed2014training} employed the bootstrapping loss to incorporate a perceptual consistency term (assigning a new label generated by the convex combination of current network prediction and the original noisy label) in the learning process. Along this research line, SELFIE \cite{song2019selfie} is known by using the co-teaching strategy to select clean samples and progressively refurbish noisy labels that most frequently predicted by previous learned models. Another typical work is Joint Optimization \cite{tanaka2018joint}, using two progressive steps to update the whole data labels and classifier weights alternatively based on the knowledge delivered in dynamic iteration of the algorithm. Besides, U-correction \cite{arazo2019unsupervised} built a two-component Beta Mixture Model (BMM) to estimate the probability of sample being mislabeled and correct noisy labels by bootstrapping loss. From the perspective of label correction, we can view all these methods as different means of generating soft labels to replace the original targets. Albeit capable of correcting noisy labels to a certain extent, the performance of these methods heavily rely on the reliability of the generated soft labels, which is depend on the accuracy of the classifier trained on the noisy dataset. When the classifier has poor performance, the false label information supplied by it will further degrade the quality of the obtained classifier. Moreover, these methods usually need to manually preset proper hyper-parameters to better fit different training data. This, on the other hand, makes them hardly generalized to variant and diverse scenarios in real cases.

To solve the above problems, in this paper we design a meta soft label corrector (MSLC), which could correct corrupted labels iteratively, from the angle of meta-learning.  Concretely, we treat the label correction procedure as a two stage optimization process. In the first stage, we generate soft labels through MSLC by utilizing the original targets and different temporal information of predictions from base model. Then we update the MSLC by gradient descent step in order to minimize the loss of clean meta data. In the second stage, we let the base learner train to fit the soft labels which generated by MSLC in the first stage. Through optimizing the two stages alternatively, it could effectively utilize the guidance of meta data and improve the performance in noisy labels. The contributions of this paper can be summarized as follows:

\begin{itemize}
\item Our method propose a meta soft label corrector which could map input label to a corrected soft label without using conventional pre-defined generating rules, and thus making the label correction process more flexible and easily adapt to complicated real dataset with different types and levels of noise.
\item Under the guidance of noise-free meta-data, our method could adaptively make use of the temporal predictions of the model to generate more accurate pseudo-labels without manually pre-set combination coefficient.
\item  Our proposed model is model agnostic and could be added on the top of any existing models at hand. Comprehensive synthetic and real experiments validate the superiority of the proposed method on robust deep learning with noisy labels. This can be finely interpreted by its obviously better noisy-clean label distinguishing capability and superior quality of the new soft labels generated by MSLC.
\end{itemize}

\section{Meta Soft Label Corrector}
\subsection{Analysis of the existing label correction methods}
For a c-class classification, let $\mathcal{X} \subset \mathbb{R}^{d}$ be the feature space, $\mathcal{Y} = \{1,2,...,c\}$ be the label space. Given training data $D =\{(x_{i},y_{i})\}_{i=1}^{N}\in(\mathcal{X}\times\mathcal{Y})^{N} $, where $x_{i}$ is the $i$-th sample with its label denoted as $y_{i}$.
Denoting the network as a function with input $x \in \mathcal{X}$ and output $f: \mathcal{X} \rightarrow \mathbb{R}^{c} $, then $f(x;w)$ is a network with $w$ representing the network parameters.
%Let $f_{j}(x;w)$ represents the $j$-th component ($j=1,\cdots,c$) of $f(x;w)$, which satisfies $\sum_{j=1}^{c}f_{j}(x;w)=1$, $f_{j}(x;w)\ge 0$.
In order to learn the model $f(x;w)$, given dataset $D$, the parameters $w$ can be optimized by a chosen loss function. %Specifically, when loss function is cross-entropy, the objective function can be define as:
%\begin{equation}%\nonumber
%  \label{E1}
%  \mathcal{L}_{CE}(D,w) = \frac{1}{N}\sum_{i=1}^{N}\sum_{j=1}^{c}y_{ij}\log(f_{j}(x_{i};w))
%\end{equation}

The label correction methods focus on how to generate more accurate pseudo-labels that could replace the original noisy ones so that increase the performance of the classifier. E.g., Reed et al. \cite{reed2014training} proposed a static hard bootstrapping loss to deal with label noise, in which the training objective for $(t+1)^{th}$ step is
\vspace{-4mm}
\begin{equation}
 \label{E2}
 \mathcal{L}(D,w) = \frac{1}{N}\sum_{i=1}^{N}l\left(f\left(x_{i};w \right), \lambda_{i} y_{i} + (1-\lambda_{i}) \hat{y}_i^{(t)} \right),
\end{equation}
where $\hat{y}^{(t)}_{i}$ is the predicted label by the classifier in the $t^{th}$ step,
$\lambda_{i} y_{i} + (1-\lambda_{i}) \hat{y}_{i}$ can be seen as a soft pseudo-label that replaces the original target $y_{i}$ with preset parameter $\lambda_{i}$, and $l(\cdot)$ is a chosen loss function.
%Reed et al. \cite{andhonglak2015training} set the $\lambda=0.8$ for the “hard” one which $\hat{y_{i}}$ in the form of one-hot, and set $\lambda=0.95$ for the "soft" one which $\hat{y_{i}}$ in the form of soft-target that could be added to 1. However, setting a small weight to model prediction $\hat{y_{i}}$ limits the ability of rectifying the original noisy target $y_{i}$ and the fixed weight for all samples can not perform well on the task of noisy labels with different types and levels of noise.
In similar formulation as Eq. (\ref{E2}), some other methods design its own strategy to generate pseudo-labels. For example, SELFIE \cite{song2019selfie} set a threshold to separate the low-loss instances as clean samples and decide which samples are corrupted according to the volatility of the predictions of samples, and then correct these by the most frequently predicted label in previous $q$ iterations.
%That means it sets $\lambda_{i} \in \{0,1\}$ according the criterion on whether the data is clean or not and sets the mode of past q epochs' predictions as $\hat{y_{ij}}$. However, SELFIE needs to know the noise ratio to set the threshold value before it applied to the noisy dataset, which limits its' application in real dataset.
Furthermore, Arazo et al. \cite{arazo2019unsupervised} learned the $\lambda_{i}$ dynamically for every sample by using a Beta-Mixture model which is an unsupervised method to group the loss of samples into two categories and choose the prediction of the $t^{th}$ step as $\hat{y}$ similar to Eq. (\ref{E2}).% \cite{reed2014training}

%they \textcolor{red}{choose the current prediction of model as $\hat{y_{i}}$}.

Different from the form of Eq. (\ref{E2}), Joint Optimization \cite{tanaka2018joint} trained their model on the original targets with cross-entropy loss in a large learning rate for several epochs, and then tried to use the network predictions to generate pseudo-labels without using the original labels. They used loss function is,
\vspace{-1.5mm}
\begin{equation}
 \label{E3}
 \mathcal{L}(D,w) = \frac{1}{N}\sum_{i=1}^{N}l \left(f\left(x_{i};w \right), \sum_{j=0}^{q-1}\hat{y}^{(t-j)}_i \right),
\end{equation}
where the pseudo-labels are the average of the predictions that from the past $q$ epochs. With a finely set  hyper-parameters $q$, it could achieve robust performance.

It can be seen that the existing label correction methods exploited a manually set mechanism for correcting labels. However, compared with specifically design to the investigated problem, it is a more difficult task to construct a unique label correction methodology that could be finely adaptable to different applications and datasets, which constitutes the main task of this work.

Moreover, these methods may cause severe error accumulation due to the low quality of the new soft labels that replaced the original ones. Bootstrap \cite{reed2014training} and U-correction\cite{arazo2019unsupervised} combined the observed label $y$ with the current prediction $\hat{y}^{(t)}$ to generate new soft labels. However, there exists significant variation in the predictions of base model especially to the samples which labels are corrupted. Joint Optimization \cite{tanaka2018joint} method used the the predictions of earlier network to alleviate this problem, but it used the new soft labels to replace all the observed targets no matter whether it's clean or not may cause the question that the clean original labels were wrongly corrected.

%However, it is a difficult task to construct a unique label correction methodology, instead of specifically design various ones, to be finely adaptable to different applications and datasets, which constitutes the main task of this work.
\subsection{Structure of the proposed MSLC}
To alleviate the aforementioned issues of the current methods, we want to build a learning framework which could generate pseudo-labels with following data-adaptive label corrector for each training step:
\begin{equation}\label{Generate Mechanism}
   \tilde{y} = g\left( y, I; \theta \right)
\end{equation}
%\begin{equation}\label{Generate Mechanism}
%   \tilde{y}_i^{(t)} = g\left(y_i, \hat{y}_i^{(t)}, \tilde{y}_i^{(t-1)}; \theta\right),
%\end{equation}
where $\tilde{y}$ is the soft pseudo-label generated by our proposed MSLC, $y$ denotes the original label, $I$ represents the side information that is helpful to make such label correction, and $\theta$ denotes hyper-parameters involved in this function. The questions are now how to specify $I$ and the function parametric format of $g$, and how to learn its involved parameters $\theta$.

%where $y_i$ is the label of the collected data, $\hat{y}_i^{(t)} = f\left(x_{i};w^{(t)} \right)$ is the prediction of the $t^{th}$ step, $\tilde{y}_i^{(t-1)}$ and $\tilde{y}_i^{(t)}$ denote the pseudo-labels generated by $(t-1)^{th}$ and $t^{th}$ steps, respectively, and $\theta$ denotes all the parameters involved in the pseudo-label generator. It should be noted that the generator contains to-be-estimated parameter, and we will show in later section how to built this generator in a deep network formulation and estimate its parameter $\theta$ automatically.

With meta soft label corrector Eq.~(\ref{Generate Mechanism}), the final training objective for $(t+1)^{th}$ step can be written as:
%\begin{equation}
% \label{E4}
% \mathcal{L}(D,w,\theta) =  -\frac{1}{N}\sum_{i=1}^{N} l\left( f\left(x_{i};w \right),  g\left(y_i, I; \theta \right)\right).
%\end{equation}
\vspace{-2mm}
\begin{equation}
 \label{E4}
 \mathcal{L}(D,w,\theta) =  \frac{1}{N}\sum_{i=1}^{N} l\left( f\left(x_{i};w \right),   \tilde{y}_i^{(t)}\right).
\end{equation}

%\begin{equation}
% \label{E4}
% \mathcal{L}(D,w,\theta) = -\frac{1}{N}\sum_{i=1}^{N} l\left( f\left(x_{i};w \right),   \tilde{y}_i^{(t)}\right) = -\frac{1}{N}\sum_{i=1}^{N} l\left( f\left(x_{i};w \right),  g\left(y_i, \hat{y}_i^{(t)}, \tilde{y}_i^{(t-1)}, \theta \right)\right).
%\end{equation}

%\subsection{Structure of the proposed MSLC}
 Synthesize these helpful experience that we analyzed previous section, we use $\hat{y}^{(t)}$ and $\tilde{y}^{(t-1)}$ Eq.~(\ref{Generate Mechanism}) as the side information for helping correcting the input label $y$ \footnote{Note that more earlier generated pseudo-labels $\tilde{y}^{(t-j)}$ for $j>1$ could be easily adopted in our method. Our experiments show that one projection $\tilde{y}^{(t-1)}$ can already guarantee a good performance. We thus easily use this simple setting, but could readily explore to use more in future. In this sense, both pseudo-label utilization manners (e.g., \cite{reed2014training} and \cite{tanaka2018joint}) as introduced in Section 3.1 can be seen as special cases of ours, but with manually preset combination coefficients instead of automatically learned ones directly from data like ours.
}, i.e.,

\vspace{-3mm}
\begin{equation}
   g\left(y, I ; \theta \right) = g\left(y, \hat{y}^{(t)}, \tilde{y}^{(t-1)};\theta\right).
\end{equation}

It is worth noting that U-correction \cite{arazo2019unsupervised} adopt an unsupervised model to learn the hyper-parameter, however, possibly due to the alternatively updating procedure of the unsupervised model and the base classifier, although it could fit well to the fixed loss distribution, it can not split noisy samples accurately in the training process(See section 3.2). To alleviate these issues, we view the label correction procedure as a meta-process and using a meta-learner to automatically correct labels. Inspired by \cite{reed2014training} and \cite{tanaka2018joint}, we easily set the corrected label to be a convex combination of $y, \hat{y}^{(t)}, \tilde{y}^{(t-1)}$. That is:
\vspace{-1.5mm}
\begin{equation}\label{Detail}
  g\left(y, \hat{y}^{(t)}, \tilde{y}^{(t-1)}; \theta\right) \!=\! \alpha (l^\alpha; \theta_\alpha)y+\left(1\!-\!\alpha(l^\alpha; \theta_\alpha)\right)\!\left( \beta(l^\beta; \theta_\beta)\tilde{y}^{(t-1)} + (1\!-\!\beta(l^\beta; \theta_\beta)) \hat{y}^{(t)} \right),
\end{equation}
\vspace{-2mm}

%\begin{multicols}{2}[\columnsep2em]
%where $\alpha(\cdot)$ and $\beta(\cdot)$ are two networks, whose outputs represent coefficients of this convex combination, with their parameters denoted as $\theta_\alpha$ and $\theta_\beta$, respectively, and thus $\theta=[\theta_\alpha, \theta_\beta]$. These two coefficient networks, with $l^\alpha = l(\hat{y}^{(t)},y)$  and $l^\beta = l(\hat{y}^{(t)},\tilde{y}^{(t-1)})$, then constitute the main parts of our proposed label corrector, which is intuitively shown in  Fig. \ref{PM}.
%where $\alpha(\cdot)$ and $\beta(\cdot)$ are two networks, whose outputs represent coefficients of this convex combination, with their parameters denoted as $\theta_\alpha$ and $\theta_\beta$, respectively, and thus $\theta=[\theta_\alpha, \theta_\beta]$. These two coefficient networks, with $l^\alpha = l(\hat{y}^{(t)},y)$  and $l^\beta = l(\hat{y}^{(t)},\tilde{y}^{(t-1)})$, then constitute the main parts of our proposed label corrector, which is intuitively shown in  Fig. \ref{PM}.
%
%\columnbreak
%
%
%\includegraphics[width=2.5in]{./fig/method/generator3.pdf}
%%\captionof{The structure of the proposed label corrector function, e.g., the Purifying Machine.} % $\tilde{y}^{(t-1)}$}
%
%\end{multicols}

\begin{minipage}[b]{0.45\linewidth}
where $\alpha(\cdot)$ and $\beta(\cdot)$ are two networks, whose outputs represent coefficients of this convex combination, with their parameters denoted as $\theta_\alpha$ and $\theta_\beta$, respectively, and thus $\theta=[\theta_\alpha, \theta_\beta]$. These two coefficient networks, with $l^\alpha = l(\hat{y}^{(t)},y)$  and $l^\beta = l(\hat{y}^{(t)},\tilde{y}^{(t-1)})$, then constitute the main parts of our proposed soft label corrector, which is intuitively shown in  Fig. \ref{PM}. Through the two networks, the input target information, i.e. $y$, $\hat{y}^{(t)}$,$\tilde{y}^{(t-1)}$ , could be combined in a convex combination to form a new soft target, $\tilde{y}^{(t)}$, which will replace the original label $y$ in the training process. $\alpha$ and $\beta$ are the output value of $\alpha(\cdot)$ and $\beta(\cdot)$ respectively.

%The $y,\hat{y}^{(t)},\tilde{y}^{(t-1)}$ denote three types information of targets, $\tilde{y}^{(t)}$ is the output of the label corrector to replace the original target $y$ in training.
\end{minipage}
\hfill
\begin{minipage}[b]{0.5\linewidth}
\includegraphics[width=3.4in, height=1.65in]{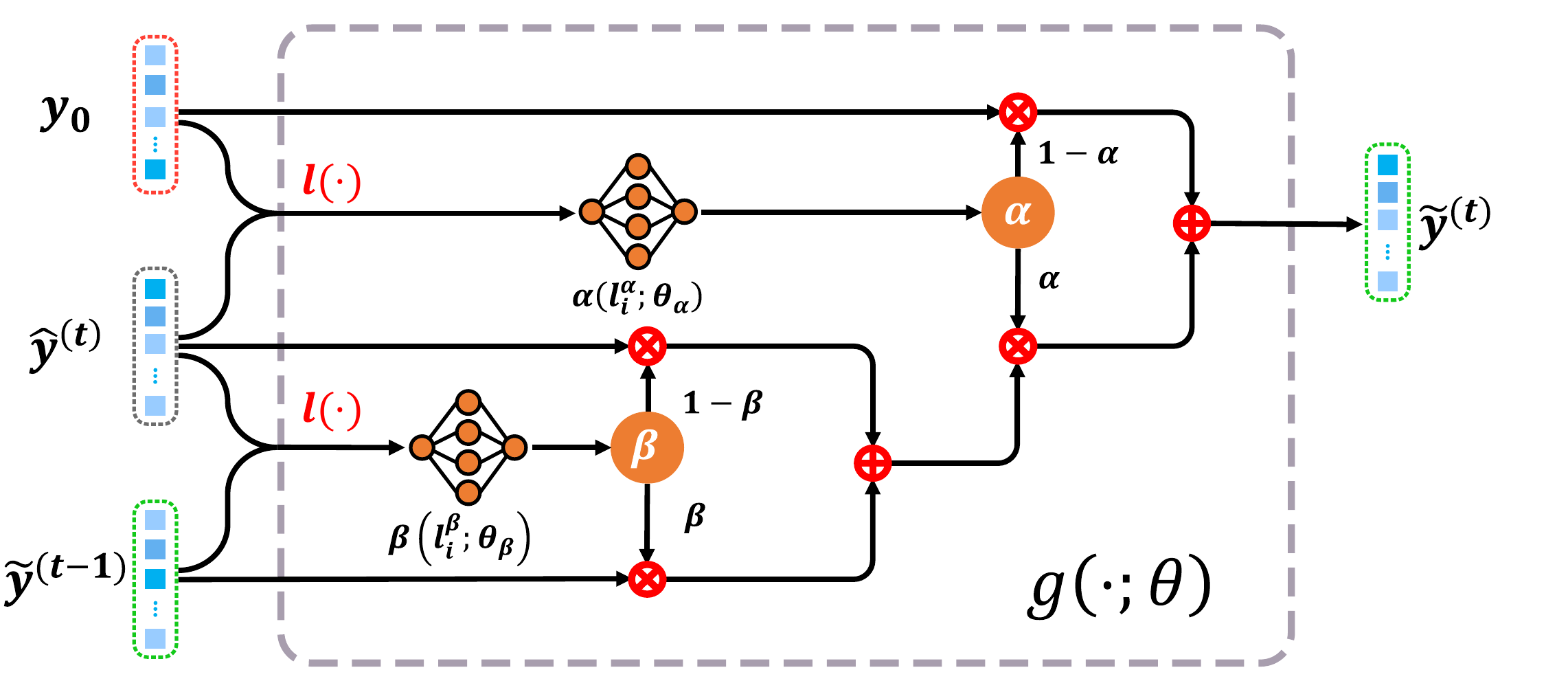}
\captionof{figure}{The Structure of MSLC \label{PM}}
%\captionof {The structure of the proposed label corrector function, e.g., the Purifying Machine.} % $\tilde{y}^{(t-1)}$}
\end{minipage}

Our proposed MSLC exploits meta-learning method and could better distinguish the noisy and clean samples than the unsupervised manner. Also we take more temporal predictions information into consideration so that the generated new soft labels are more accurate and could effectively prevent severe error accumulation.

%where $\alpha(\cdot)$ and $\beta(\cdot)$ are two networks, whose outputs represent coefficients of this convex combination, with their parameters denoted as $\theta_\alpha$ and $\theta_\beta$, respectively, and thus $\theta=[\theta_\alpha, \theta_\beta]$. These two coefficient networks, with $l^\alpha = l(\hat{y}^{(t)},y)$  and $l^\beta = l(\hat{y}^{(t)},\tilde{y}^{(t-1)})$, then constitute the main parts of our proposed label corrector, which is intuitively shown in  Fig. \ref{PM}.

%\begin{figure}[t]
%
%  \centering
%
%    \includegraphics[width=2.5in]{./fig/method/generator3.pdf}
%
%  \caption{The structure of the proposed label corrector function, e.g., the Purifying Machine.} % $\tilde{y}^{(t-1)}$}
%  \label{PM}
%  %\label{fig:subfig} %% label for entire figure
%\end{figure}

\subsection{Training with meta dataset}

We then introduce how to learn hyper-parameter $\theta$ for the MSLC Eq.~(\ref{Detail}). We readily employ a meta-data driven learning regime as used in \cite{shu2019meta}, which exploits a small but noise free dataset (i.e., meta data) for learning the hyper-parameter for training samples. The meta dataset contains the meta-knowledge of underlying label distribution of clean samples, it is thus rationally to exploit it as a sound guide to help estimate $\theta$ for our task. Such data can be seen as the conventional validation data (but with high quality), with much smaller size than those used for training, and thus feasible to be pre-collected. In this work, we denoted meta dataset as,
\begin{equation}\label{Meat}
  \mathcal{D}_{meta}=\{ (x_{i}^{meta},y_{i}^{meta})\}_{i=1}^{M},
\end{equation}
where $M $($M\ll N$) is the number of data samples in  meta dataset.
By utilizing the meta dataset, we can then design the entire training framework for the noise label correction model Eq.~(\ref{E4}).

Specifically, we formulate the following bi-level minimization problem:
 \vspace{-0.2mm}
\begin{equation}
 \label{E8}
 w^{*}(\theta)= \arg\min_{w} \mathcal{L}(D,w;\theta)\qquad \theta^{*}= \arg\min_{\theta} \mathcal{L}_{meta}(w^*(\theta)),
\end{equation}
%\begin{equation}
% \label{E9}
% \theta^{*}= \arg\min_{\theta} \mathcal{L}_{meta}(w^*(\theta)),
%\end{equation}
where $\mathcal{L}_{meta}(w)=\frac{1}{M}\sum_{i=1}^{M}l\left(f\left(x_i^{meta};w\right),y_i^{meta}\right)$ is the meta loss on meta dataset. After achieving $\theta^*$, we can then get the soft label corrector, which incline to ameliorate noisy labels to be correct ones, and further improve the quality of the trained classifier.

\begin{figure}[t]

  \centering

    \includegraphics[width=5.5in, height = 1.22in]{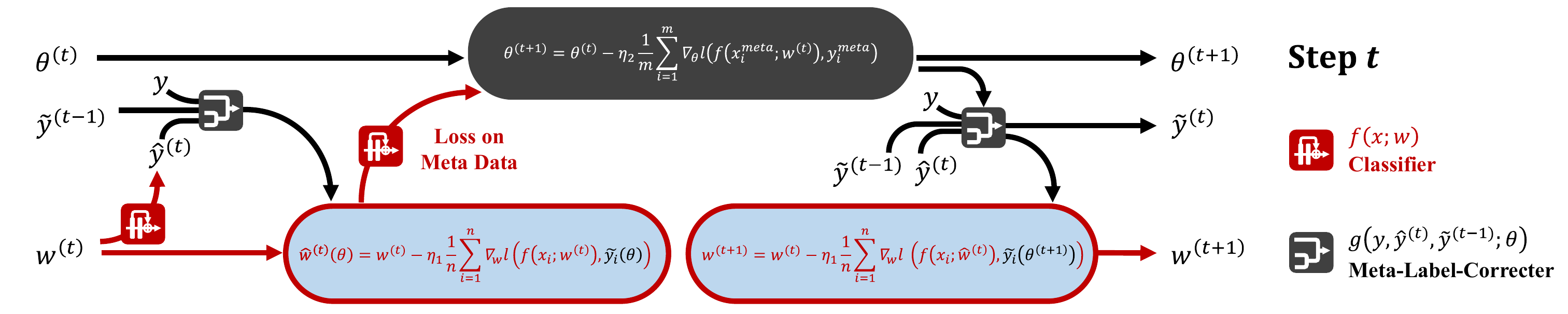}

  \caption{Main flowchart of the proposed MSLC }
  \label{PMAlgorithm}
  %\label{fig:subfig} %% label for entire figure
\end{figure}

Optimizing the parameters $w$ and hyper-parameters $\theta$ requires two nested loop of optimization Eq.~(\ref{E8}), which tends to be computationally inefficient \cite{franceschi2018bilevel}. We thus exploit SGD technique to speedup the algorithm by approximately solving the problem in a mini-batch updating manner \cite{shu2019meta,finn2017model} to jointly ameliorating $\theta$ and $w$. The algorithm flowchart is shown in Fig. \ref{PMAlgorithm}.

\begin{algorithm}[t]\label{algo}
	\vspace{0mm}
	\renewcommand{\algorithmicrequire}{\textbf{Input:}}
	\renewcommand{\algorithmicensure}{\textbf{Output:}}
	\caption{The Learning Algorithm of Meta Label Corrector}
	\label{alg1}
	\begin{algorithmic}[1]  \small
		% \STATE {\bfseries Input:} Training data $\mathcal{D}$, meta set $\mathcal{D}^{(m)}$, batch size $n,n^{(m)}$, max iterations $T$.
		%   \STATE {\bfseries Output:} Student model parameters $w^{(T)}$
		\REQUIRE  Training data $D$, meta data $D_{meta}$, batch size $n,m$, MaxEpoch $T$.
		\ENSURE  Classifier network parameter $w^{(T)}$ %, hyperparameter $\alpha$, $\beta$.
		%\REPEAT
		\STATE Initialize classifier parameter $w^{(0)}$ and Meta-Learner parameter  $\theta^{(0)}$.
		\FOR{$t=1$ {\bfseries to} $T$}
		\STATE $\{x,y\} \leftarrow$ SampleMiniBatch($D,n$).
		\STATE $\{x^{(m)},y^{(m)}\} \leftarrow$ SampleMiniBatch($D_{meta},m$).
		\STATE Update $\theta^{(t+1)}$ by Eq. (\ref{E11}).
		\STATE Update $w^{(t+1)}$ by Eq. (\ref{E12}).
        \STATE Update $\hat{y}_i^{(t+1)}, \tilde{y}_i^{(t)}$ by the current classifier with parameter $ w^{(t+1)}$.
        %\STATE Update $\tilde{y}_i^{(t-1)}$ by the current classifier with parameter $ w^{(t+1)}$.
		\ENDFOR
		%\UNTIL{$noChange$ is $true$}
	\end{algorithmic}
\vspace{-1mm}
\end{algorithm}

%\subsection{Algorithm for Learning $\lambda$ and $\alpha$}

The algorithm includes mainly following steps. Firstly, denote the mini-batch training samples as $\{(x_{i},y_{i})\}_{i=1}^{n}$,
and then the training loss becomes $\frac{1}{n}\sum_{i=1}^{n} l\left( f\left(x_{i};w \right),  g\left(y_i,  \hat{y}_i^{(t)}, \tilde{y}_i^{(t-1)}; \theta \right)\right)$. We can then deduce the formulate of one-step $w$ updating with respect to $\theta$ as
\begin{equation}
 \label{E10}
 \hat{w}(\theta) = w^{(t)} -\eta_1\frac{1}{n}\sum_{i=1}^{n}\left.\nabla_{w}l\left(f(x_i;w),g\left(y_i, \hat{y}_i^{(t)}, \tilde{y}_i^{(t-1)}, \theta \right)\right)\right|_{w^{(t)}},
\end{equation}
where $\eta_1$ is the learning rate.
Then, with current mini-batch meta data samples $\{ (x_{i}^{meta},y_{i}^{meta})\}_{i=1}^{m}$, we can perform one step updating for solving $
\min_{\theta} \frac{1}{m}\sum_{i=1}^{m}l\left(f\left(x_i^{meta};w\right),y_i^{meta}\right)$, that is
\begin{equation}
 \label{E11}
\theta^{(t+1)} = \theta^{(t)} - \eta_{2}\frac{1}{m}\sum_{i=1}^{m}\left.\nabla_{\theta}l\left(f(x_i^{meta};\hat{w}(\theta)), y_i^{meta}\right)\right|_{w^{(t)}},
\end{equation}
 After we achieve $\theta^{(t+1)}$, we can calculated the pseudo label by Eq.~(\ref{Generate Mechanism}), and update $w$, that is
\begin{equation}
 \label{E12}
 w^{(t+1)} = w^{(t)} -\eta_1\frac{1}{n}\sum_{i=1}^{n}\left.\nabla_{w}l\left(f(x_i;w),g\left(y_i, \hat{y}_i^{(t)}, \tilde{y}_i^{(t-1)}; \theta^{(t+1)} \right)\right)\right|_{w^{(t)}}.
\end{equation}
The predicted pseudo-labels $\hat{y}_i^{(t+1)}, \tilde{y}_i^{(t)}$ can then be updated by using the current classifier with parameter $ w^{(t+1)}$.
The entire algorithm is then summarized in Algorithm 1.%, which we call it Purify Machine.%LP-Net (label purify network).

%\subsection{Implementation details}
%In the proposed method, we chose cross-entropy [???] as loss function for all of our experiments, and set $eta_1$ as...
\vspace{-3.5mm}
\section{Experimental Results}
\vspace{-2mm}
To evaluate the capability of the proposed method, we implement experiments on CIFAR-10, CIFAR-100 \cite{krizhevsky2009learning} under different types and levels of noise, as well as a real-word large-scale noisy dataset Clothing1M \cite{xiao2015learning}. Both CIFAR-10 and CIFAR-100 contain 50k training images and 10k test images of size 32 $\times$ 32. For CIFAR-10/100, we use two types of label noise: symmetric and asymmetric. \textbf{Symmetric:} We follow \cite{zhang2016understanding,tanaka2018joint} for label noise addition, which generates label corruptions by flipping labels of a given proportion of training samples to one of the other class labels uniformly (the true label could be randomly maintained). \textbf{Asymmetric:} We use the setting in \cite{yao2019safeguarded}, which designs to mimic the structure of real-world label noise. Concretely, we set a probability $r$ to disturb the label to its similar class, e.g., truck $\rightarrow$ automobile, bird $\rightarrow$ airplane, deer $\rightarrow$ horse, cat $\rightarrow$ dog. For CIFAR-100, a similar $r$ is set but the label flip only happens in each super-class as described in \cite{hendrycks2018using}.

\textbf{Baselines.} The compared methods include: \textbf{Fine-tuning}, which finetunes the result of Cross-Entropy on the meta-data to further enhance its performance. \textbf{GCE} \cite{zhang2018generalized}, which employs a robust loss combining the benefits of both CE loss and mean absolute error loss against label noise. \textbf{GLC} \cite{hendrycks2018using}, which estimates the noise transition matrix by using a small clean label dataset. \textbf{MW-Net} \cite{shu2019meta}, which uses a MLP net to learn the weighting function. \textbf{Bootstrap} \cite{reed2014training}, which deals with label noise by adding a perceptual term to the standard CE loss. \textbf{Joint Optimization} \cite{tanaka2018joint}, which updates the label and model at the same time by using the pseudo-labels it generated.  \textbf{U-correction} \cite{arazo2019unsupervised}, which models sample loss with BMM and applied MixUp. For fair comparison, we only compare its proposed method without mixup augmentation.

%\textbf{SELFIE} \cite{song2019selfie}, which splits the clean and noisy samples and then correct the corrupted ones;

\textbf{Experiment Details.} We use ResNet-34 \cite{he2016deep} as classifier network for all baseline experiments in Table 1 .We use two multi-layer perception(MLP) with 100 hidden layers as the network structure of $\alpha(\cdot)$ and $\beta(\cdot)$ respectively. In the proposed method, we chose cross-entropy as loss function, we began to correct labels at 80th epoch (i.e. there is an initial warm-up).%, the other settings are shown in Appendix.
%Table 1 followed by \cite{patrini2017making} \cite{xia2019anchor}
% except for SELFIE \cite{song2019selfie}, which uses DenseNet (L=25, k=12)
%In the proposed method, we chose cross-entropy as loss function, and set $\eta_{1}=0.001$,$\eta_{2}=0.001$ for all of our experiments. We use SGD with a momentum of 0.9, a weight decay of $5 \times 10^{-4}$, and batchsize of 100. We began to correct labels at 80th epoch (i.e. there is an initial warm-up), and the learning rate is 0.1 which is divided  by 10 after 80 and 100 epochs for a total of 120 epochs.

\begin{table*}[t] \vspace{-4mm}
	\caption{Test accuracy (\%) of all competing methods on CIFAR-10 and CIFAR-100 under Symmetric  noise and Asymmetric  noise with different noise levels. The best results are highlighted in \textbf{bold}.}
    \label{table111}
	\centering
	\setlength{\tabcolsep}{0.9mm}
	\begin{scriptsize}
		\begin{tabular}{c|c|c|c|c|c|c|c|c|c|c|c|c|c|c} %\multicolumn{2}{c}{\multirow{2}{*}{Multi-col-row}}
			\toprule
			\multicolumn{3}{c|}{\multirow{1}{*}{Noise-type}}   & \multicolumn{8}{c|}{\multirow{1}{*}{Symmetric Noise}}    &  \multicolumn{4}{c}{\multirow{1}{*}{Asymmetric Noise}}  \\ \cline{1-15}
			\multicolumn{3}{c|}{\multirow{1}{*}{Dataset}}   & \multicolumn{4}{c|}{\multirow{1}{*}{CIFAR-10}}   & \multicolumn{4}{c|}{\multirow{1}{*}{CIFAR-100}}  &  \multicolumn{2}{c|}{\multirow{1}{*}{CIFAR-10}} & \multicolumn{2}{c}{\multirow{1}{*}{CIFAR-100}}  \\ \cline{1-15}

            \multicolumn{2}{c}{\multirow{1}{*}{Method $\backslash$ Noise ratio $\gamma$ }} &
            & \multicolumn{1}{|c|}{0.2}  &           0.4    &             0.6 &            0.8  & \multicolumn{1}{|c|}{0.2} & 0.4 &  0.6 &  0.8  &\multicolumn{1}{|c|}{0.2} & 0.4 &\multicolumn{1}{|c|}{0.2} & 0.4\\ \cline{1-15}

            \multicolumn{2}{c}{\multirow{2}{*}{Cross-Entropy}}    &\multicolumn{1}{c|}{Best}   & 90.22 & 87.33 & 83.2 & 54.79& 68.03& 61.18& 46.43& 17.91 &92.85   & 90.22 & 69.05 & 65.14 \\
            \multicolumn{1}{c}{} &\multicolumn{1}{c}{}   &\multicolumn{1}{c|}{Last}    & 86.33 & 79.61 & 72.99 & 54.26& 63.67& 46.92& 30.96& 8.29 &91.29   & 87.23 & 63.68 & 50.10  \\ \cline{1-15}

            \multicolumn{2}{c}{\multirow{2}{*}{Fine-tuning}}    &\multicolumn{1}{c|}{Best}   & 91.17 & 87.34 & 83.75 & 56.28& 67.81& 62.55& 50.82& 19.05 &93.11   & 91.04 & 69.55 & 65.75 \\
            \multicolumn{1}{c}{} &\multicolumn{1}{c}{}   &\multicolumn{1}{c|}{Last}   & 88.27 & 82.16 & 79.36 & 54.82& 63.97& 51.14& 38.22& 18.86 &92.35   & 89.49 & 66.43 & 55.08\\ \cline{1-15}

            \multicolumn{2}{c}{\multirow{2}{*}{GCE\cite{zhang2018generalized}}}    &\multicolumn{1}{c|}{Best}   & 90.27 & 88.50 & 83.70 & 57.27& 71.36& 63.39& 58.06& 16.51 &90.11  & 85.24 & 69.56 & 57.50\\
            \multicolumn{1}{c}{} &\multicolumn{1}{c}{}   &\multicolumn{1}{c|}{Last}   & 90.15 & 88.01 & 82.87 & 57.22& 71.02& 52.15& 45.31& 15.71 &89.33   & 82.04 & 66.36 & 56.81 \\ \cline{1-15}

            \multicolumn{2}{c}{\multirow{2}{*}{GLC\cite{hendrycks2018using}}}    &\multicolumn{1}{c|}{Best}   & 91.43 & 88.52 & 84.08 & 64.21& 69.30& 63.24& 56.12& 18.59 &92.46   & 91.74 & 71.40 & 67.73  \\
            \multicolumn{1}{c}{} &\multicolumn{1}{c}{}   &\multicolumn{1}{c|}{Last}   & 90.13 & 87.04 & 82.63 & 62.19& 66.62& 59.03& 51.96& 8.08 &92.41   & 91.02 & 70.01 & 66.68 \\ \cline{1-15}

            \multicolumn{2}{c}{\multirow{2}{*}{MW-Net\cite{shu2019meta}}}    &\multicolumn{1}{c|}{Best}    & 91.48 & 87.34 & 81.98 &65.88& 69.79& 65.44& 55.42& 19.62 &93.44  & 91.64 & 67.54 & 60.24  \\
            \multicolumn{1}{c}{} &\multicolumn{1}{c}{}   &\multicolumn{1}{c|}{Last}    & 90.11 & 86.42 & 81.62 & 64.78& 68.37& 64.81& 55.04& 19.20 &91.95   & 90.88 & 66.71 & 59.53 \\ \cline{1-15}

            \multicolumn{2}{c}{\multirow{2}{*}{Bootstrap\cite{reed2014training}}}    &\multicolumn{1}{c|}{Best}   & 91.46 & 88.75 & 84.03 & 63.80& 69.79& 63.73& 57.20& 17.63 &93.08   & 91.18 & 70.93 & 67.82  \\
            \multicolumn{1}{c}{} &\multicolumn{1}{c}{}   &\multicolumn{1}{c|}{Last}   & 88.00 & 83.57 & 78.69 & 63.41& 63.00& 47.08& 35.86& 17.04 &91.02   & 85.59 & 63.46 & 49.18 \\ \cline{1-15}

            \multicolumn{2}{c}{\multirow{2}{*}{Joint Optimization\cite{tanaka2018joint}}}    &\multicolumn{1}{c|}{Best} & 90.85 & 90.27 & 86.49 & 66.39& 63.84& 59.82& 49.13& 18.95 &93.39   & 91.43 & 66.90 & 64.82 \\
            \multicolumn{1}{c}{} &\multicolumn{1}{c}{}   &\multicolumn{1}{c|}{Last}   & 89.77 & 88.58 & 85.57 & 65.92& 60.10& 56.85& 47.68& 17.38 &92.12   & 90.20 & 66.69 & 59.31  \\ \cline{1-15}

            \multicolumn{2}{c}{\multirow{2}{*}{U-correction\cite{arazo2019unsupervised}}}    &\multicolumn{1}{c|}{Best}  & 92.05 & 89.07 & 85.64 & 68.23& 68.37& 62.37& 55.19& 17.10 &91.85   & 90.34 & 67.71 & 66.75\\
            \multicolumn{1}{c}{} &\multicolumn{1}{c}{}   &\multicolumn{1}{c|}{Last}   & 90.21 & 85.45 & 83.15 & 64.78& 67.42& 55.40& 55.04& 9.33  &90.92   &84.31  & 63.82 & 60.64 \\ \cline{1-15}

            \multicolumn{2}{c}{\multirow{2}{*}{Ours}}    &\multicolumn{1}{c|}{Best}    & \textbf{93.46} & \textbf{91.42} & \textbf{87.39} & \textbf{69.87}& \textbf{72.51}& \textbf{68.98}& \textbf{60.81}& \textbf{24.32} &\textbf{94.39}   & \textbf{92.81} & \textbf{72.66} & \textbf{70.51}\\
            \multicolumn{1}{c}{} &\multicolumn{1}{c}{}   &\multicolumn{1}{c|}{Last}    &\textbf{93.38} & \textbf{91.21} & \textbf{87.25} & \textbf{68.88}& \textbf{72.02}& \textbf{68.70}& \textbf{60.25}&\textbf{20.53} &\textbf{94.11} & \textbf{92.48} & \textbf{70.20} & \textbf{69.24} \\ \cline{1-15}

			%\bottomrule
		\end{tabular} %\vspace{-5mm}
	\end{scriptsize}
\end{table*}
\vspace{-1mm}
\subsection{Comparison with State-of-the-Art Methods} %and Analysis
\vspace{-1mm}
Table \ref{table111} shows the results of all competing methods on CIFAR-10 and CIFAR-100 under symmetric and asymmetric noise as aforementioned. To compare different methods in more detail, we report both the best test accuracy and the averaged test accuracy over the last 5 epochs.  It can be observed that our method gets the best performance across the both datasets and all noise rates. Specifically, even under relatively high noise ratios (E.g. $\gamma=0.8$ on CIFAR-10 with sym-noise), our algorithm has competitive classification accuracy ($69.87\%$). It worth noted that U-correction achieved best accuracy of $68.23\%$ that is comparable with, while its accuracy decreases in the later training as $64.78\%$ probably due to its error accumulation. This indicating that our proposed meta soft label corrector has better convergence under the guidance of meta data in the training process. It also can be seen that MW-Net has poor performance in asymmetric condition, that might because all classes share one weighting function in the method, which is unreasonable when noise is asymmetric. Comparatively, our proposed MSLC has a higher degree of freedom and thus performs much better with asymmetric noise.

\begin{figure}[h]
  \vspace{-3mm}
  \centering
  \subfigure[Cifar100-Sym-Noise40$\%$]{
    %\label{fig:subfig:a} %% label for first subfigure
    \includegraphics[width=4.2cm]{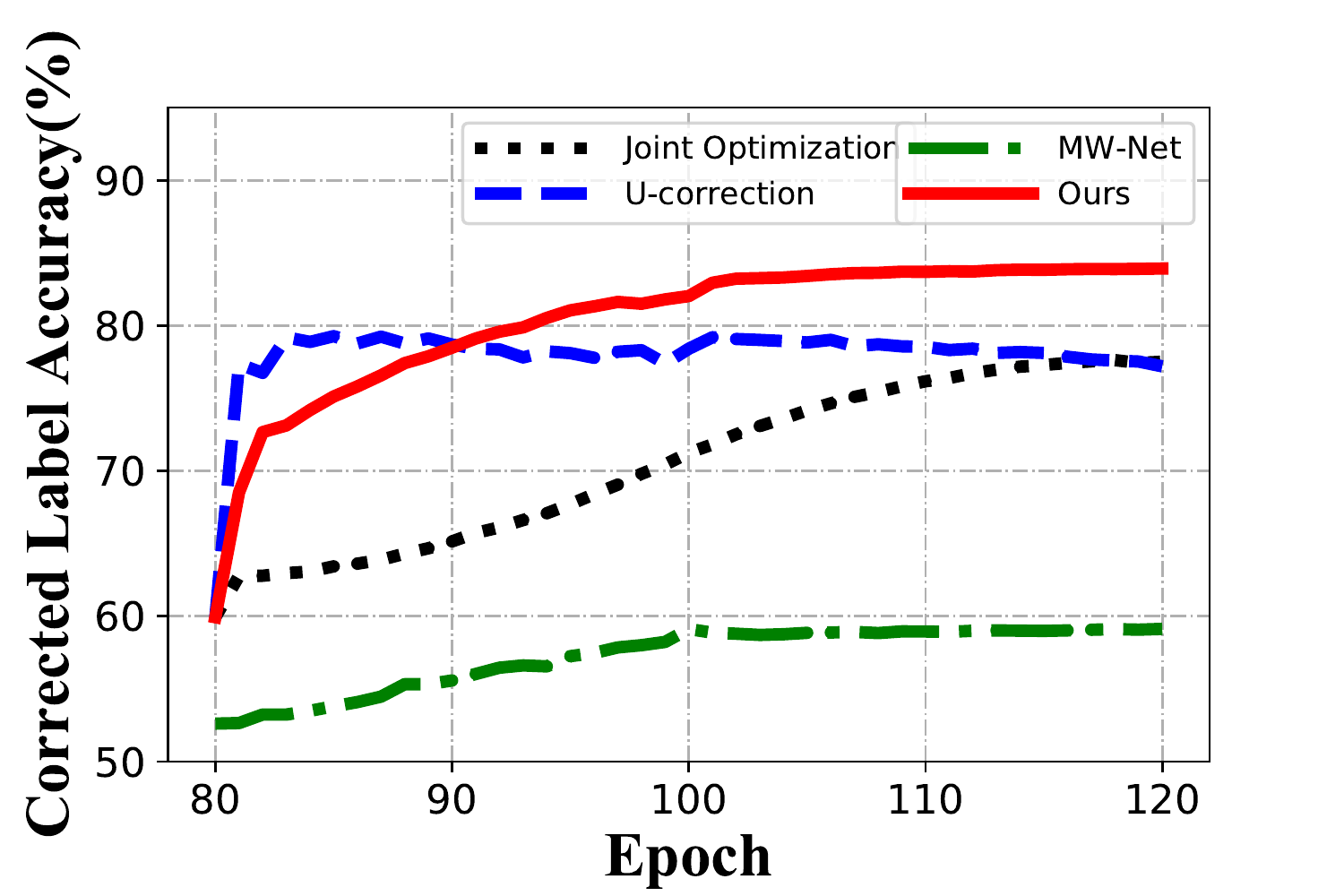}
  }
  \subfigure[Cifar100-Sym-Noise60$\%$]{
    %\label{fig:subfig:b} %% label for second subfigure
    \includegraphics[width=4.2cm]{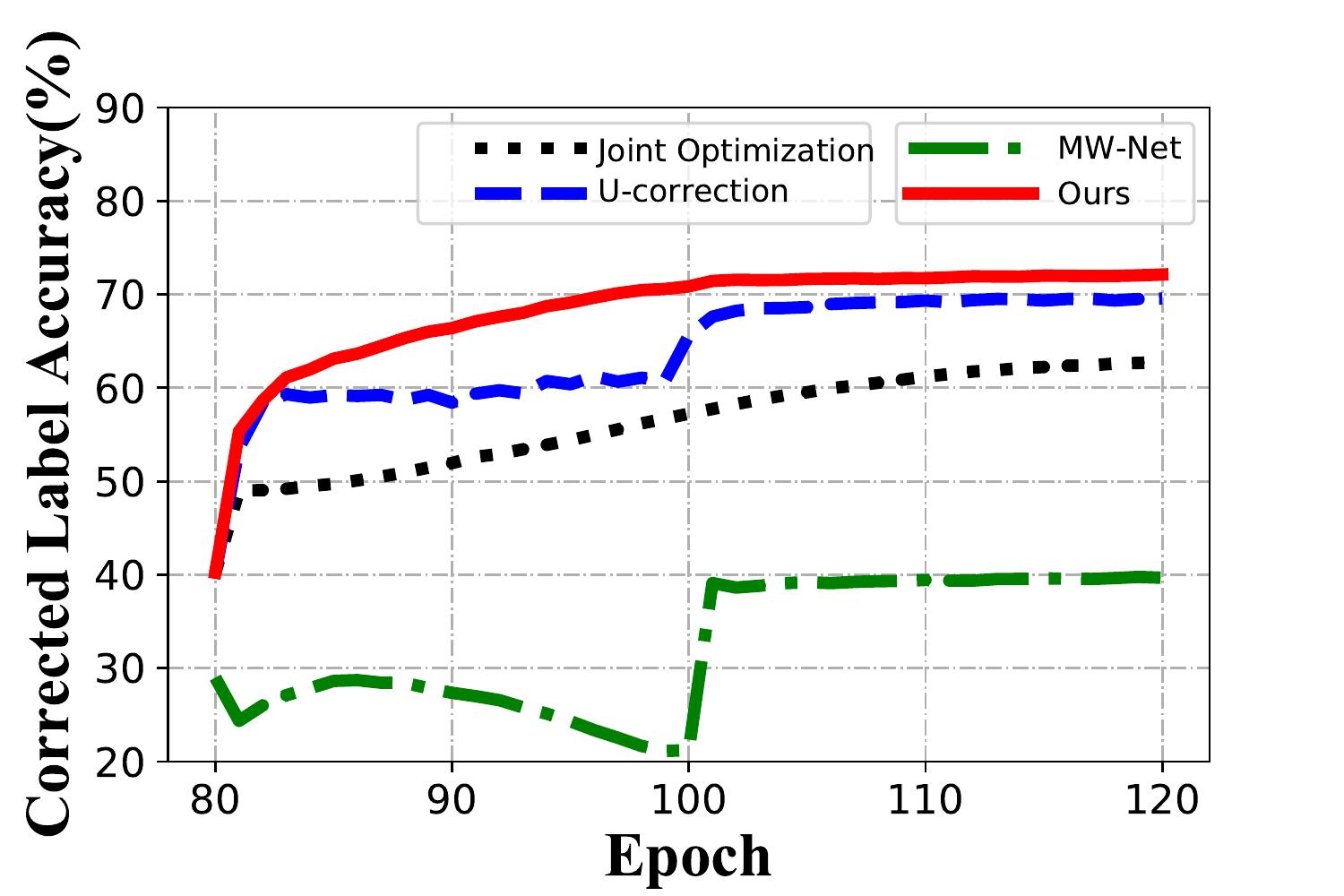}
  }
  \subfigure[Cifar100-Asy-Noise40$\%$]{
    %\label{fig:subfig:b} %% label for second subfigure
    \includegraphics[width=4.2cm]{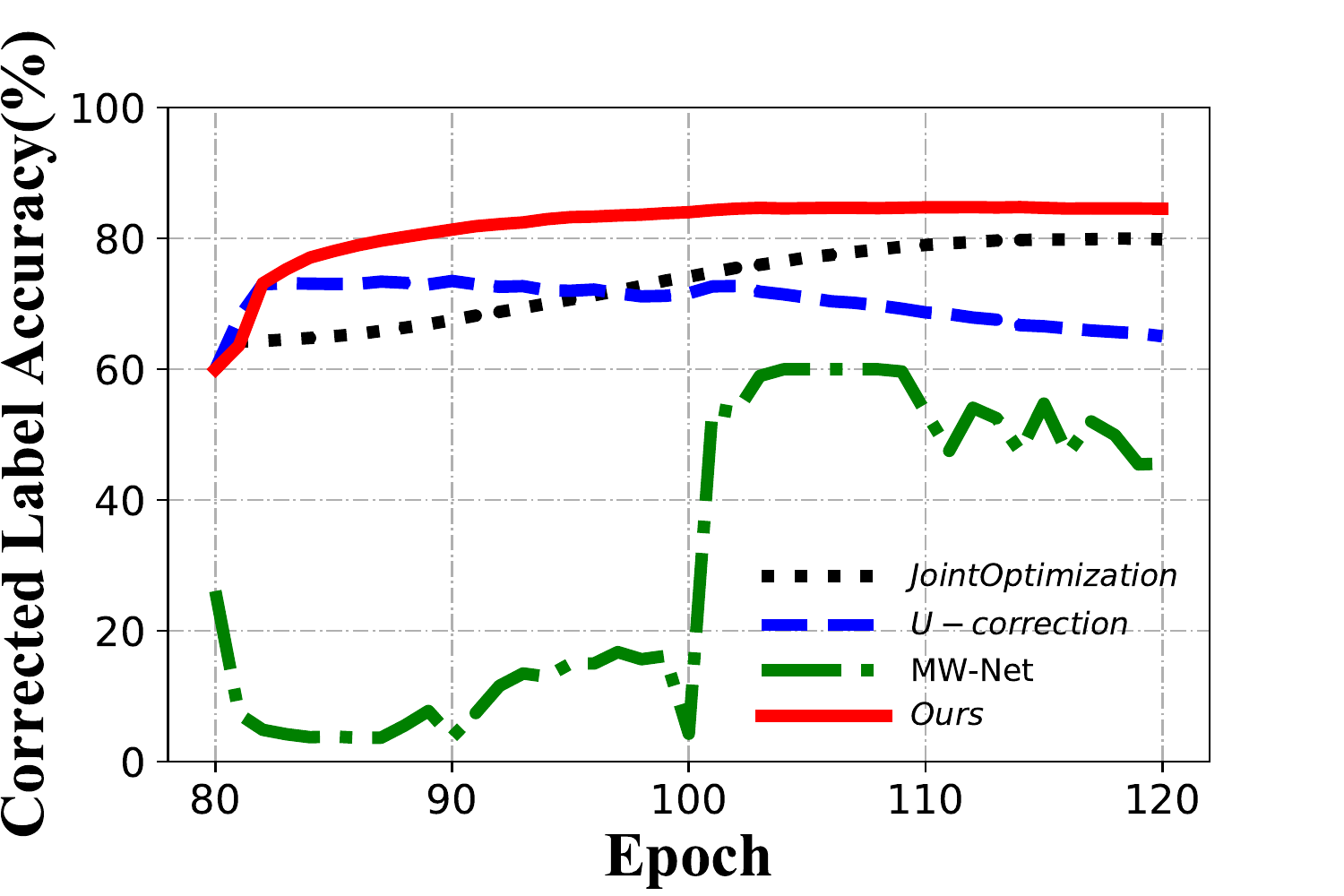}
  }
  \caption{The corrected label accuracy on different noise types and noise ratios. (a) shows the accuracy of $40\%$ symmetric noise on Cifar100, (b) shows the accuracy of $60\%$ symmetric noise on Cifar100, (c) shows the accuracy of $40\%$ asymmetric noise on Cifar100 }
  \label{LA}
  %\label{fig:subfig} %% label for entire figure
\end{figure}%
\vspace{-2mm}
%\begin{figure}[t]
%  \centering
%  \subfigure[Asym-Noise40$\%$]{
%    %\label{fig:subfig:a} %% label for first subfigure
%    \includegraphics[width=1.31in]{./fig/confusion_matrix_new/A40BC1.pdf}
%    \includegraphics[width=1.31in]{./fig/confusion_matrix_new/A40AC1.pdf}
%  }
%%  \subfigure[After Correction]{
%%    %\label{fig:subfig:b} %% label for second subfigure
%%    \includegraphics[width=1.2in]{./fig/confusion_matrix_new/A40AC.pdf}
%%  }
%  \subfigure[Sym-Noise40$\%$]{
%    %\label{fig:subfig:b} %% label for second subfigure
%    \includegraphics[width=1.31in]{./fig/confusion_matrix_new/U40BC1.pdf}
%    \includegraphics[width=1.31in]{./fig/confusion_matrix_new/U40AC1.pdf}
%  }
%%  \subfigure[After Correction]{
%%    %\label{fig:subfig:b} %% label for second subfigure
%%    \includegraphics[width=1.2in]{./fig/confusion_matrix_new/U40AC.pdf}
%%  }
%  \caption{The comparison of confusion matrices between before and after correction on CIFAR-10 with (a) asymmetric noise $40\%$ and (b) symmetric noise $40\%$. }
%  \label{CM}
%  %\label{fig:subfig} %% label for entire figure
%\end{figure}

Fig.\ref{LA} plots the corrected label accuracy, which used the hard form of pseudo-labels Eq.~(\ref{Generate Mechanism}) compared with the ground truth. As can be seen in Fig. \ref{LA}, the corrected labels generated by our method are the most accurate. The accuracy of MW-Net always below the value of the proportion of clean samples, since it intrinsically tries to select the clean samples while ignores the corrupted ones by its weighting mechanism. From Fig. \ref{LA} (a)(c), we could see that the corrected label accuracy of the U-correction are slightly decrease, it might be caused by its massive false correction\footnote[2]{This will be further analysis in the section 3.2}. Moreover, although the accuracy of JointOptimization increase all the time, its performance is limited by the strategy that only use the pseudo-labels to replace all the targets, which has the risk of corrupting the original clean labels\footnotemark[2].

%which will seen in Fig.\ref{explain}.

%For better comparison, we set these methods began to update labels from the 80th epoch except for MW-Net, where the setting have been verified by us that has no significant effect on the experimental results of these methods.
%\begin{figure}[h]
%  \vspace{-4mm}
%  \centering
%  \subfigure[The output weight of $\alpha(\cdot)$ ]{
%    %\label{fig:subfig:a} %% label for first subfigure
%    \includegraphics[width=4.2cm]{./fig/clean_noise_split/lambda_vis1.pdf}
%  }
%  \subfigure[Accuracy in clean/noisy samples corresponding to  Fig.\ref{LA} (a)]{
%    %\label{fig:subfig:b} %% label for second subfigure
%    \includegraphics[width=4.2cm]{./fig/clean_noise_split/unif_40_clean.pdf}
%    \includegraphics[width=4.2cm]{./fig/clean_noise_split/unif_40_corrupt.pdf}
%  }
%%  \subfigure[Cifar100-Asy-Noise40$\%$]{
%%    %\label{fig:subfig:b} %% label for second subfigure
%%    \includegraphics[width=4.2cm]{./fig/clean_noise_split/asy_40_corrupt.pdf}
%%  }
%  \caption{The Analysis of the proposed method on CIFAR-100 with Symmetric $40\%$ noise. (a) denotes the output weight of $\alpha(\cdot)$ on clean/noise samples, (b) shows the corrected label accuracy on clean/noisy data which split by the whole dataset according to the ground-truth.}
%  \label{explain} \vspace{-4mm}
%\end{figure}

Tabel \ref{clothing1M} are the results on real noisy dataset Clothing1M, which consists of 1 million clothing images belonging to 14 classes from online shopping websites e.g. T-shirt, Shirt, Knitwear and additional smaller sets with clean labels for validation(14K) and testing(10K). Since the labels are generated by using surrounding texts of the images provided by the sellers, they thus contain many error labels. From Table \ref{clothing1M}, it can be observed that the proposed method achieves the best performance, which indicates our meta soft label corrector could be applied to real complicated dataset. % The experimental details are shown in Appendix.

\begin{table*}[h] \vspace{-1mm}
	\caption{Test accuracy ($\%$) of different models on real-world noisy dataset Clothing1M.}\label{table1} \vspace{1mm}
    \label{clothing1M}
	\centering
	\setlength{\tabcolsep}{5mm}
	\begin{scriptsize}
		\begin{tabular}{c|c|c|c|c|c|c|c|c|c|c|c} %\multicolumn{2}{c}{\multirow{2}{*}{Multi-col-row}}
			\toprule
			\multicolumn{1}{c|}{\multirow{1}{*}{$\#$}}  & \multicolumn{3}{c|}{\multirow{1}{*}{Method}}& \multicolumn{2}{c|}{\multirow{1}{*}{Accuracy}} & \multicolumn{1}{c|}{\multirow{1}{*}{$\#$}}  & \multicolumn{3}{c|}{\multirow{1}{*}{Method}}& \multicolumn{2}{c}{\multirow{1}{*}{Accuracy}}   \\

			\hline \hline \multicolumn{1}{c|}{\multirow{1}{*}{1}}  & \multicolumn{3}{c|}{\multirow{1}{*}{Cross Entropy}}& \multicolumn{2}{c|}{\multirow{1}{*}{68.94}} & \multicolumn{1}{c|}{\multirow{1}{*}{4}}  & \multicolumn{3}{c|}{\multirow{1}{*}{Joint Optimization\cite{tanaka2018joint}}}& \multicolumn{2}{c}{\multirow{1}{*}{72.23}} \\

			\multicolumn{1}{c|}{\multirow{1}{*}{2}}  & \multicolumn{3}{c|}{\multirow{1}{*}{Bootstrapping\cite{reed2014training}}}& \multicolumn{2}{c|}{\multirow{1}{*}{69.12}} & \multicolumn{1}{c|}{\multirow{1}{*}{5}}  & \multicolumn{3}{c|}{\multirow{1}{*}{MW-Net\cite{shu2019meta}}}& \multicolumn{2}{c}{\multirow{1}{*}{73.72}} \\

			\multicolumn{1}{c|}{\multirow{1}{*}{3}}  & \multicolumn{3}{c|}{\multirow{1}{*}{U-correction\cite{arazo2019unsupervised}}}& \multicolumn{2}{c|}{\multirow{1}{*}{71.00}} & \multicolumn{1}{c|}{\multirow{1}{*}{6}}  & \multicolumn{3}{c|}{\multirow{1}{*}{Ours}}& \multicolumn{2}{c}{\multirow{1}{*}{\textbf{74.02}}} \\ \cline{1-12}

			%\bottomrule
		\end{tabular} \vspace{-4mm}
	\end{scriptsize}
\end{table*}

\subsection{Analysis of the proposed MSLC}
\vspace{-1mm}
\begin{figure}[t]
  \centering
  \subfigure[Asym-Noise40$\%$]{
    %\label{fig:subfig:a} %% label for first subfigure
    \includegraphics[width=1.31in]{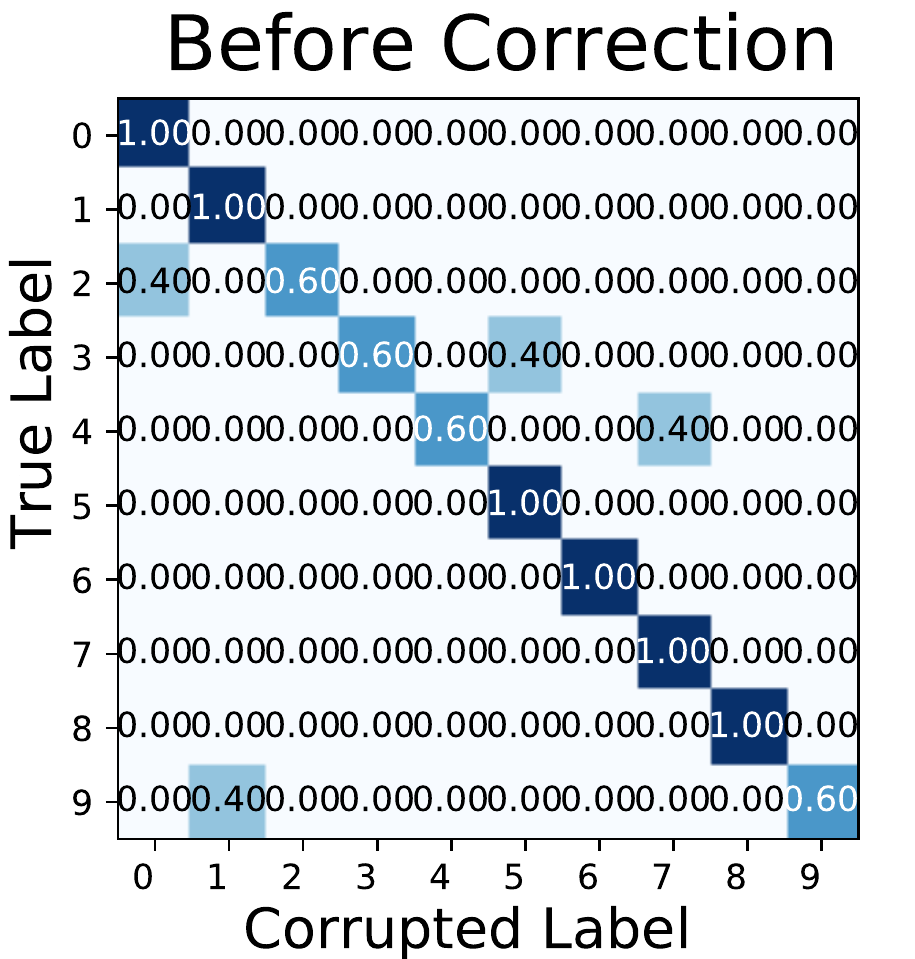}
    \includegraphics[width=1.31in]{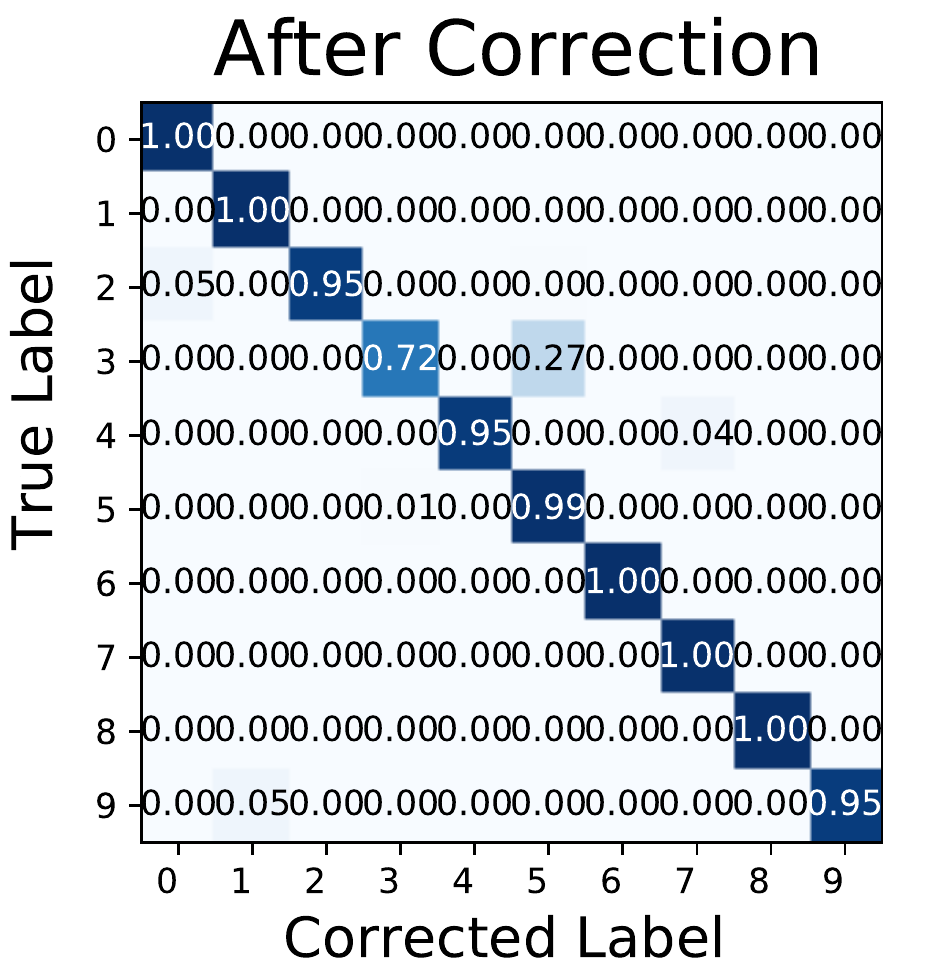}
  }
%  \subfigure[After Correction]{
%    %\label{fig:subfig:b} %% label for second subfigure
%    \includegraphics[width=1.2in]{./fig/confusion_matrix_new/A40AC.pdf}
%  }
  \subfigure[Sym-Noise40$\%$]{
    %\label{fig:subfig:b} %% label for second subfigure
    \includegraphics[width=1.31in]{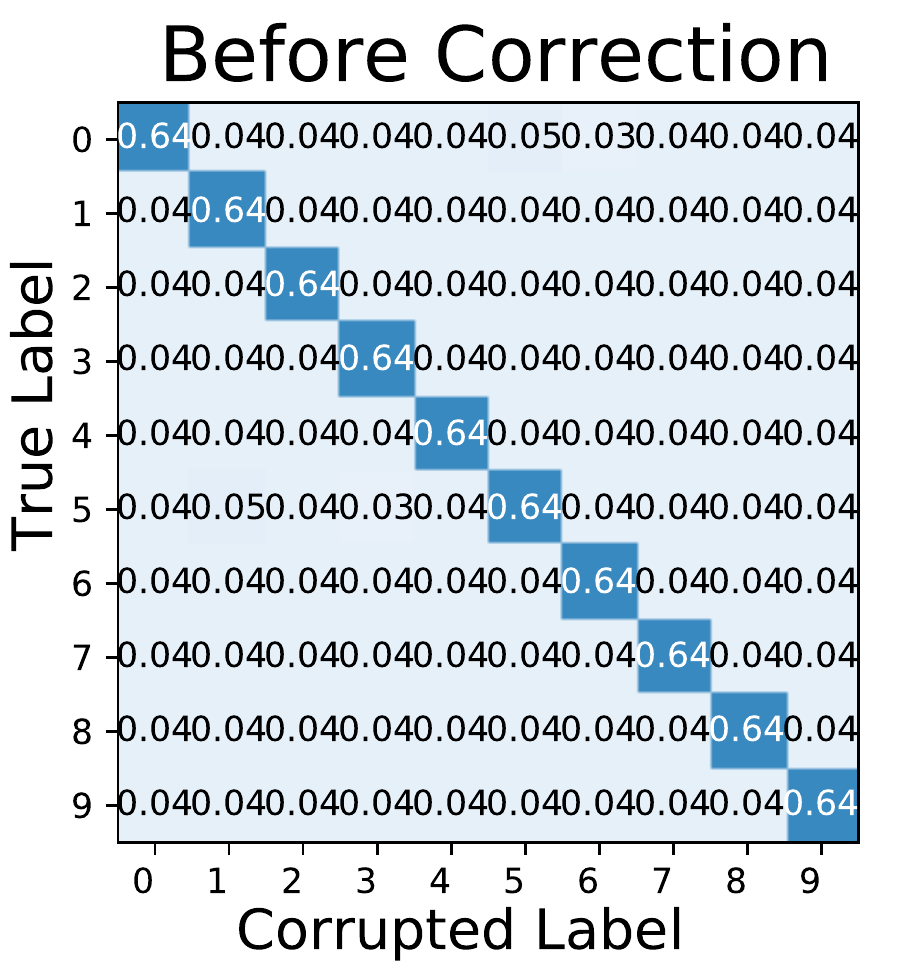}
    \includegraphics[width=1.31in]{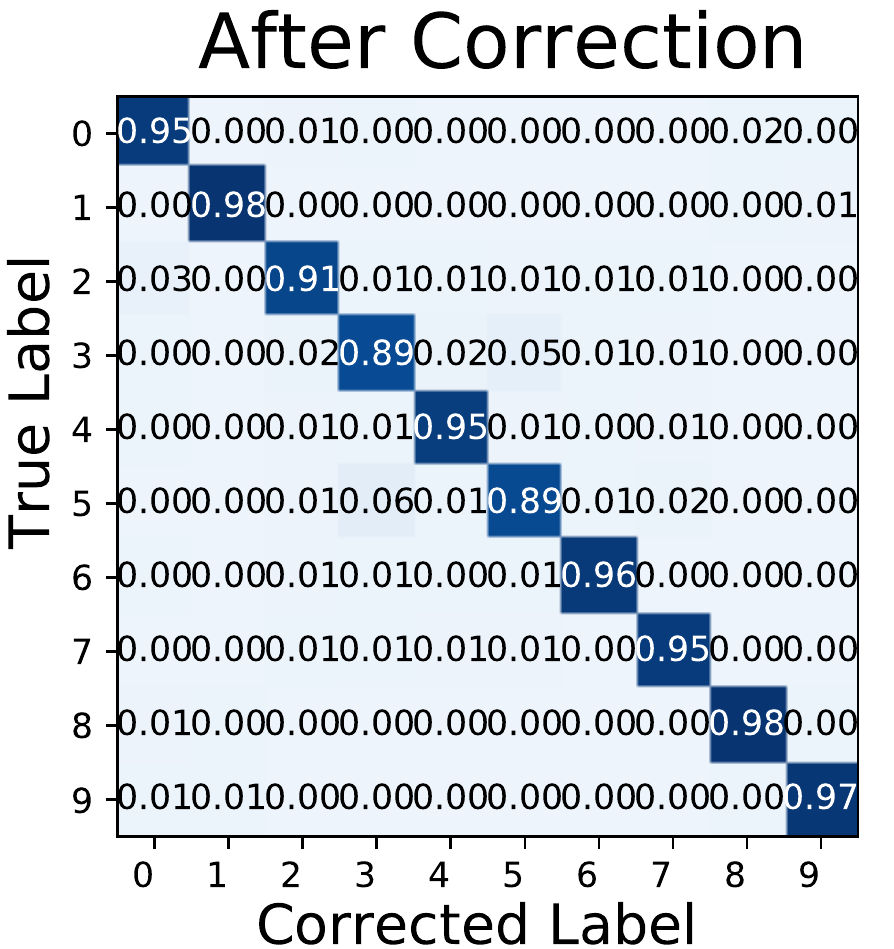}
  }
%  \subfigure[After Correction]{
%    %\label{fig:subfig:b} %% label for second subfigure
%    \includegraphics[width=1.2in]{./fig/confusion_matrix_new/U40AC.pdf}
%  }
  \caption{The comparison of confusion matrices between before and after correction on CIFAR-10 with (a) asymmetric noise $40\%$ and (b) symmetric noise $40\%$. }
  \label{CM}
  %\label{fig:subfig} %% label for entire figure
\end{figure}

\begin{figure}[t]
  \vspace{-4mm}
  \centering
  \subfigure[The output weight of $\alpha(\cdot)$ ]{
    %\label{fig:subfig:a} %% label for first subfigure
    \includegraphics[width=4.2cm]{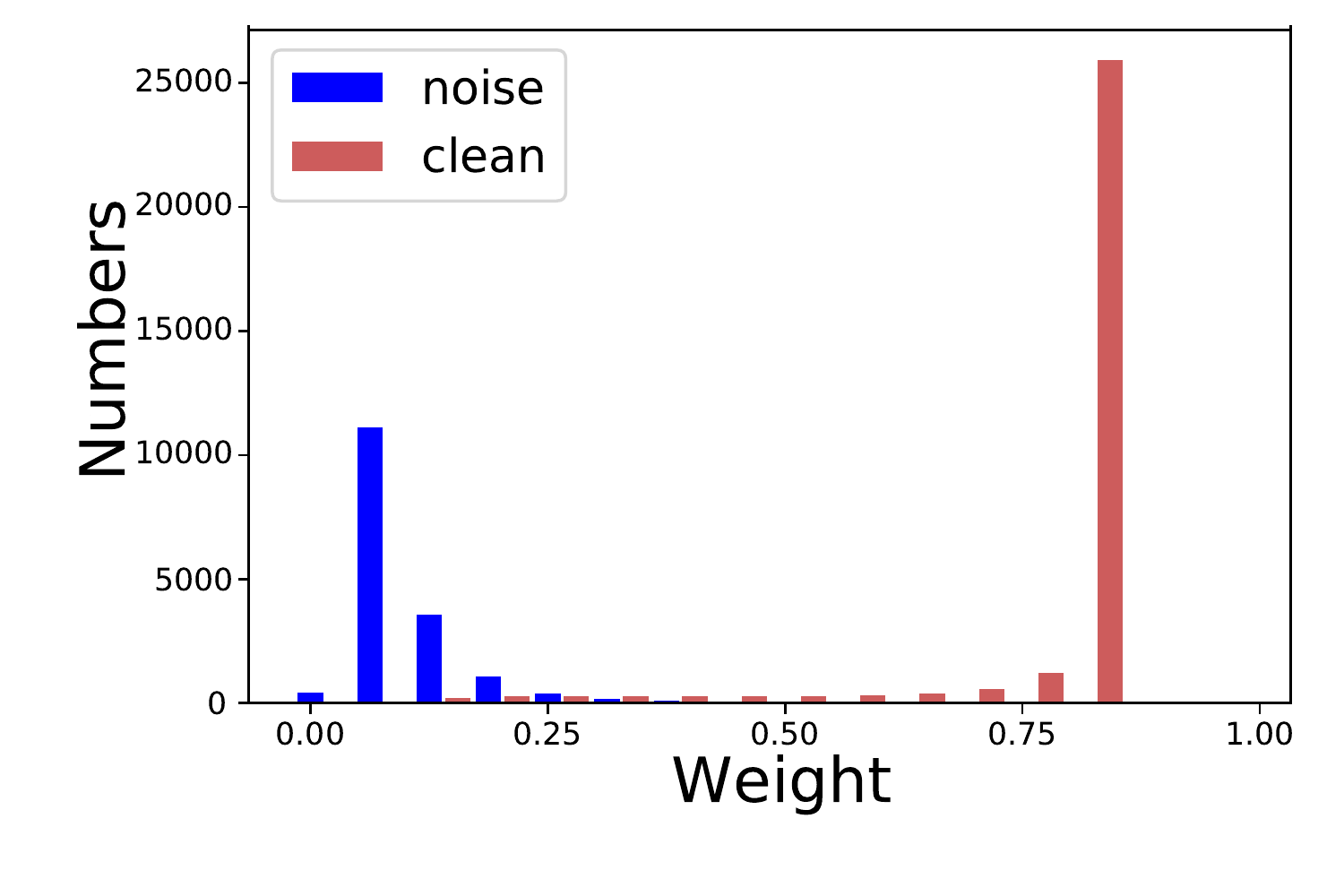}
  }
  \subfigure[Accuracy in clean/noisy samples corresponding to  Fig.\ref{LA} (a)]{
    %\label{fig:subfig:b} %% label for second subfigure
    \includegraphics[width=4.2cm]{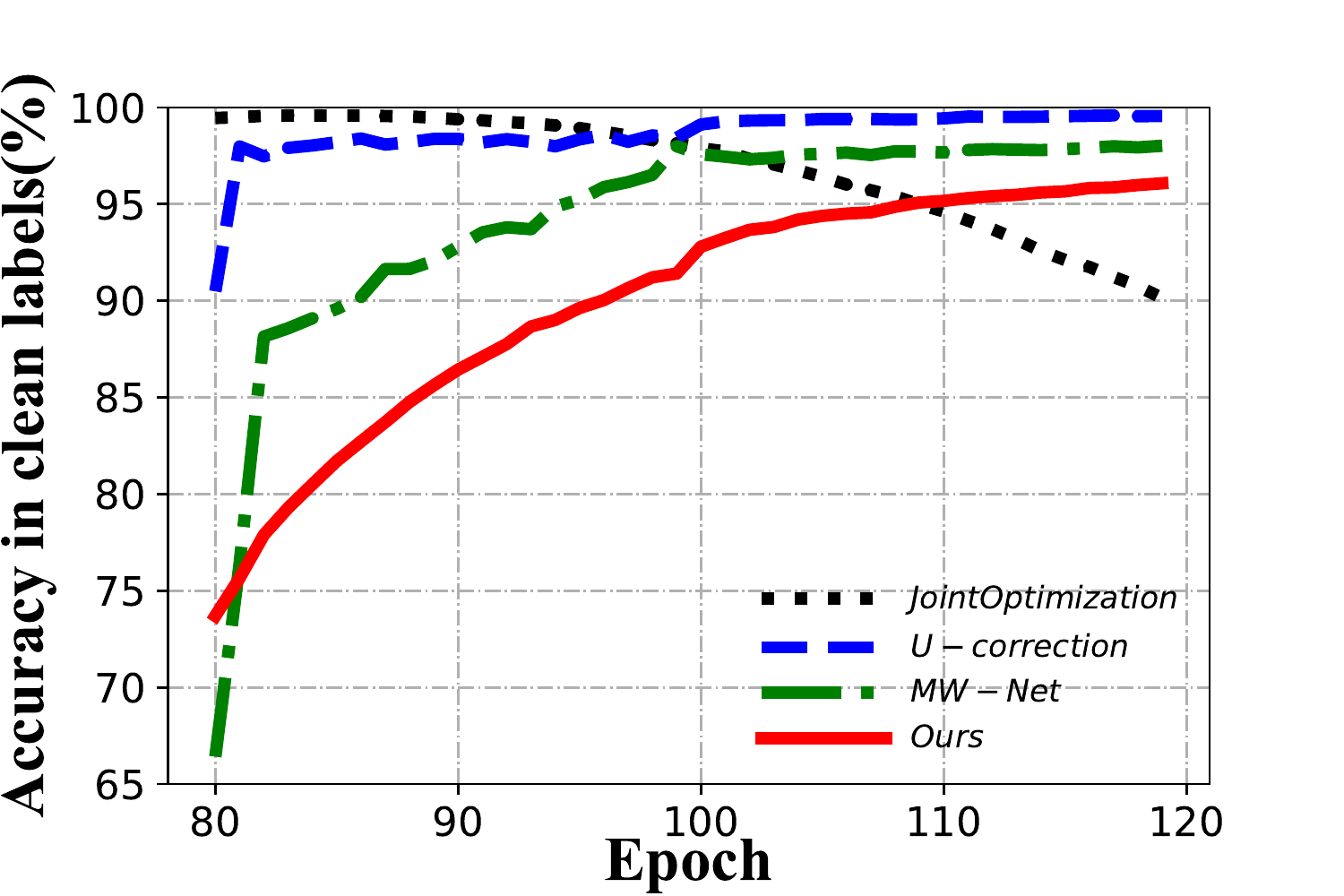}
    \includegraphics[width=4.2cm]{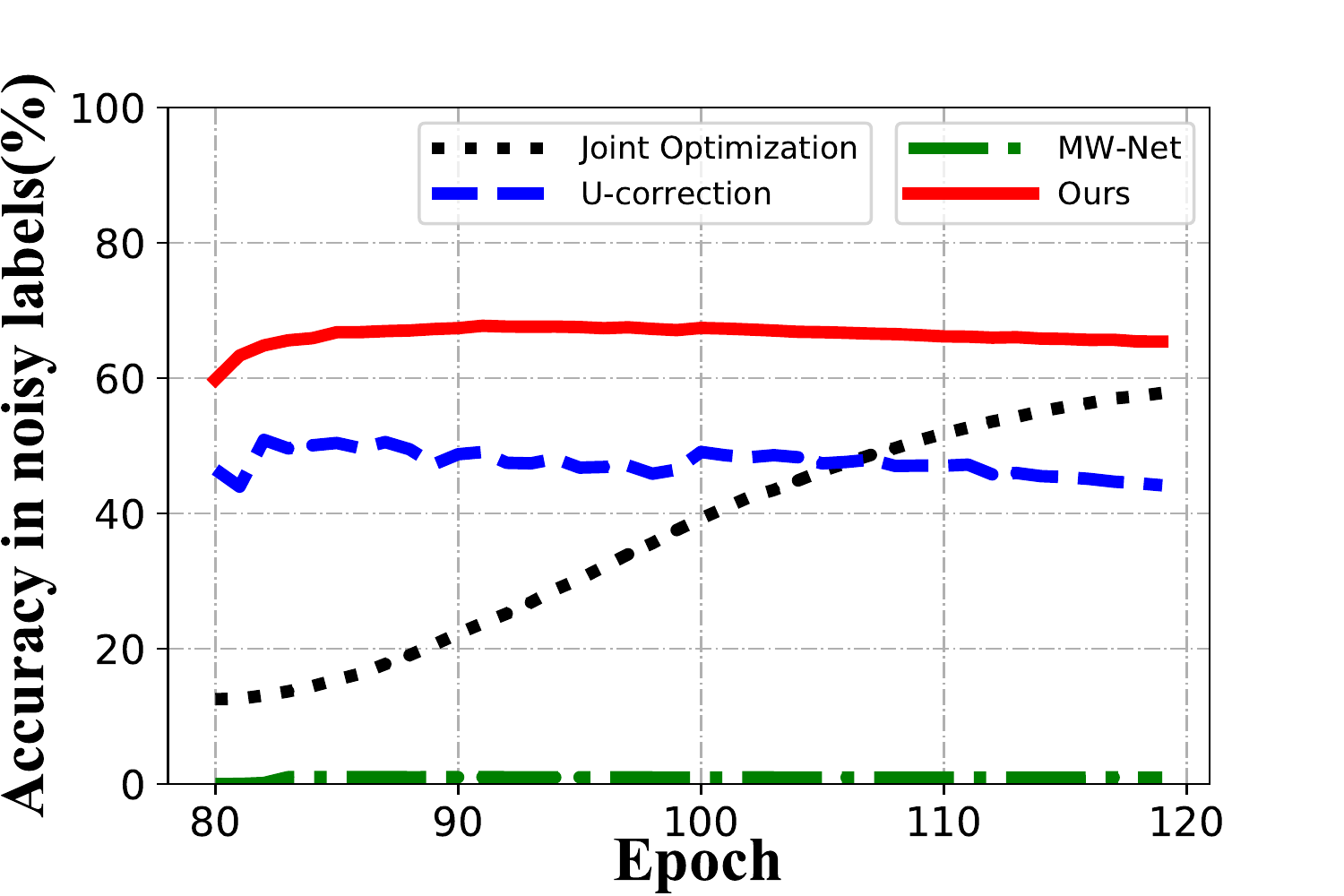}
  }
%  \subfigure[Cifar100-Asy-Noise40$\%$]{
%    %\label{fig:subfig:b} %% label for second subfigure
%    \includegraphics[width=4.2cm]{./fig/clean_noise_split/asy_40_corrupt.pdf}
%  }
  \caption{The Analysis of the proposed method on CIFAR-100 with Symmetric $40\%$ noise. (a) denotes the output weight of $\alpha(\cdot)$ on clean/noise samples, (b) shows the corrected label accuracy on clean/noisy data which split by the whole dataset according to the ground-truth.}
  \label{explain} \vspace{-4mm}
\end{figure}

\begin{table*}[h]
	\caption{The test accuracy(\%) of ablation study on CIFAR-10/100 under 40\% of sym-noise. Mean accuracy  over 3 repetitions are reported.}
    \label{ablation} %\vspace{1mm}
	\centering
	\setlength{\tabcolsep}{1.5mm}
	\begin{scriptsize}
		\begin{tabular}{c|c|c|c|c|c|c|c|c|c|c|c|c|c|c} %\multicolumn{2}{c}{\multirow{2}{*}{Multi-col-row}}
			\toprule
			\multicolumn{3}{c|}{\multirow{1}{*}{Dataset}}   & \multicolumn{6}{c|}{\multirow{1}{*}{CIFAR-10}}    &  \multicolumn{6}{c}{\multirow{1}{*}{CIFAR-100}}  \\ \cline{1-15}

			\multicolumn{3}{c|}{\multirow{1}{*}{$\beta$}}   & \multicolumn{1}{|c|}{0} & 0.2 &           0.4    &             0.6 &            0.8 & Ours & \multicolumn{1}{|c|}{0} & 0.2 & 0.4 &  0.6 &  0.8 & Ours \\ \cline{1-15}

            \multicolumn{2}{c}{\multirow{2}{*}{Accuracy}}    &\multicolumn{1}{c|}{Best} &89.84   & 90.49 & 91.04 & 90.34 & 89.46
 & \textbf{91.27}& 67.42& 68.52& 68.25& 67.13& 67.08& \textbf{68.84}  \\
            \multicolumn{1}{c}{} &\multicolumn{1}{c}{}   &\multicolumn{1}{c|}{Last} &89.46   & 90.19 & 90.91 & 89.64 & 89.20
& \textbf{91.11}& 66.93& 68.06& 67.83& 66.61& 66.24&  \textbf{68.35}   \\ \cline{1-15}
%            \multicolumn{2}{c}{\multirow{1}{*}{Method $\backslash$ Noise ratio }} &
%            & \multicolumn{1}{|c|}{0} & 0.2 &           0.4    &             0.6 &            0.8  & \multicolumn{1}{|c|}{0} & 0.2 & 0.4 &  0.6 &  0.8  &\multicolumn{1}{|c|}{0.2} & 0.4 &\multicolumn{1}{|c|}{0.2} & 0.4\\ \cline{1-17}
            \multicolumn{3}{c|}{\multirow{1}{*}{Corrected Label Accuracy}}  &92.23   & 93.36 & 94.24 & 93.44 & 91.94 & \textbf{94.52}& 81.47& 83.29 & 83.04& 81.28& 81.24 & \textbf{83.98}  \\ \cline{1-15}

			%\bottomrule
		\end{tabular} \vspace{-1mm}
	\end{scriptsize}
\end{table*}
Fig.\ref{CM} shows the confusion matrices of our method under symmetric and asymmetric noise on CIFAR-10. The left column of Fig.\ref{CM} (a) and (b) is the noise transition matrix, which is the guideline for generating the synthesized noisy datasets. And the right column is the matrix after corrected by our proposed method, which x-axis denotes the hard form corrected labels. By comparing the left and right column of Fig.\ref{CM} (a) and (b), we could see that the probability of most diagonal terms exceeds $0.95$ after correction. That could indicate the high correction accuracy of our proposed MSLC.

Fig.\ref{explain} demonstrates the output weights of $\alpha(\cdot)$ and the corrected labels accuracy on clean and noisy samples, respectively. From Fig.\ref{explain} (a), we could see that the weights of clean and noisy samples are significantly different, that means our meta soft label corrector inclines to choose the original clean labels and prones to use other target information when the original labels are noisy. Fig.\ref{explain} (b) explains that our method could greatly correct the noise samples while retaining the original clean samples. It is worth noting that U-correction retains more than 99\% of clean samples, however, we through experiments have found the reason is that it inclines to treat most of the samples as clean ones in the training process, which limits its ability to correct noise samples, as show in right column of Fig.\ref{explain}(b).  As for JointOptimization, we could see that its training process corrupted the original clean labels from the left column of (b), since it used prediction targets replaced all original labels without considering if they are clean or not.

For further analysis the effectiveness of the network $\beta(\cdot)$, we compared it learned hyper-parameters ($\beta$) with a set of different manually set values on CIFAR-10 and CIFAR-100. It can be observed from Table \ref{ablation} that the performance is worst when the $\beta$ is set to 0, which means directly choose the predictions of current model could not accurately correct the original labels.  On the other hand, we can find that the best manually set $\beta$ changes when the dataset is different. Specifically, for CIFAR-10, the best test accuracy is 91.04$\%$ corresponding to $\beta=0.4$ case, while for CIFAR-100, the best is 68.52$\%$ corresponding to $\beta=0.2$. Compared with the way of setting the hyperparameter manually, our algorithm could learn it more flexibly and achieves the best performance in both test accuracy and the corrected label accuracy.

\vspace{-3mm}
\section{Related Work}
\vspace{-3mm}
%\subsection{Robust learning with noisy label}
%Currently, various methods are proposed to deal with learning with noisy labels, which can be roughly divided into two categories: sample selection and label correction.

\textbf{Sample Selection:} The main idea of this approach is to filter out clean samples from data and train the learner only on these selected ones. Some methods along this line designed their specific selective strategies. For example, Decouple \cite{malach2017decoupling} utilized  two networks to select samples with different label predictions and then used them to update. Similarly, Co-teaching \cite{han2018co} also used two networks, but chose small-loss samples as clean ones for each network. Other methods tend to select clean samples by assigning weights to losses of all training samples, and iteratively update these weights based on the loss values during the training process. A typical method is SPL (Self-paced learning), which set smaller weights to samples with larger loss since they are more possible to be noisy samples \cite{kumar2010self,jiang2014easy,zhao2015self}. Very recently, inspired by the idea of meta-learning, some advanced sample reweighting methods have been raised. Typically, MentorNet \cite{jiang2018mentornet} pre-trained an additional teacher network with clean samples to guide the training process. Ren et al. \cite{ren2018learning} used a small set of validation data to training procedure and re-weight the backward losses of the mini-batch samples such that the updated gradient minimized the losses of those validation data. These methods usually have a more complex weighting scheme, which makes them able to deal with more general data bias and select the clean samples more accurately. In these methods, however, most noisy data useful for learning visual representations \cite{pathak2017learning,gidaris2018unsupervised} are discard from training, making them leaving large room for further performance improvement.

%The main idea of this approach is to assign weights to losses of all training samples, and iteratively update these weights based on the loss values during the training process \cite{kumar2010self,jiang2014easy,zhao2015self}. Decouple \cite{malach2017decoupling} used two networks to select samples which predictions of them are different in different network, while Co-teaching \cite{han2018co} chose its small-loss samples as clean samples for each network. Other sample selection methods mainly adopts sample re-weighting schemes by imposing weights on samples based on their reliability. Tradition methods include SPL which set smaller weights to samples with larger loss since they are more possible to be noisy samples compared with small loss one. Very recently, inspired by the idea of meta-learning, some advanced sample reweighting methods have been raised. For example, MentorNet \cite{jiang2018mentornet} pre-trained an additional teacher network with clean samples of which their labels are probably correct. Ren et al. \cite{ren2018learning} draws a small set of validation data into training procedure and re-weight the backward losses of the mini-batch samples such that the updated gradient minimized the losses of those validation data. These methods usually have much more complex weighting scheme, which makes them able to deal with more general data bias and select the clean samples more accurately, however, most noisy data that are useful for learning visual representations \cite{pathak2017learning,gidaris2018unsupervised} are discard in these approaches.

\textbf{Label Correction:} The traditional label correction approach aims to correct noisy labels to true ones through an additional inference step, such as conditional random fields \cite{vahdat2017toward}, knowledge graphs \cite{li2017learning} or directed graphical models \cite{xiao2015learning}. Recently, transition matrix approach assumes that there exists a probabilities matrix that most probably flip the true labels into ``noise" ones. There exist mainly two approaches to estimate the noise transition matrix. One is to train the classifier by pre-estimating noise transition matrix with the anchor point prior assumption. The other approach is to jointly estimate the noise transition matrix and the classifier parameters in a unified framework without employing anchor points \cite{sukhbaatar2014training,jindal2016learning,goldberger2016training,xia2019anchor}.  Besides this, some other methods exploit the predictions of network to rectify labels. For example, Joint Optimization \cite{tanaka2018joint} optimizes the parameters and updates the labels at the same time by using average prediction results of the network. SELFIE \cite{song2019selfie} used the co-teaching strategy to select clean samples and progressively refurbish noisy samples by using the most frequently predicted labels of previous learned model. Arazo et al. \cite{arazo2019unsupervised} proposed a two-component Beta Mixture Model to define whether the data is corrupted or not, and then correct them by introducing the bootstrapping loss.% We will introduce more details on the typical methods along this line in Section 3.1 for better clarifying the motivation of this work.

\vspace{-4mm}
\section{Conclusion}
\vspace{-4mm}
Combining with meta-learning, we proposed a novel label correction method that could adaptively ameliorating corrupted labels for robust deep learning when the training data is corrupted. Compared with current label correction methods that use a pre-fixed generation mechanism and require manually set hyper-parameters, our method is able to do this task in a flexible automatic and adaptive data-driven manner. Experimental results show consistent superiority of our method in datasets with different types and levels of noise. In the future study, we will try to construct a new structure of meta soft label corrector, which input is not only the loss information, so that its well-trained model could transfer to other datasets under different noise level.

\nocite{*}
\bibliographystyle{unsrt}
\bibliography{ref}

\begin{appendix}
\begin{figure}[h]
  \centering
  \subfigure[Cifar100-Asy-Noise40$\%$]{
    %\label{fig:subfig:a} %% label for first subfigure
    \includegraphics[width=2.1in]{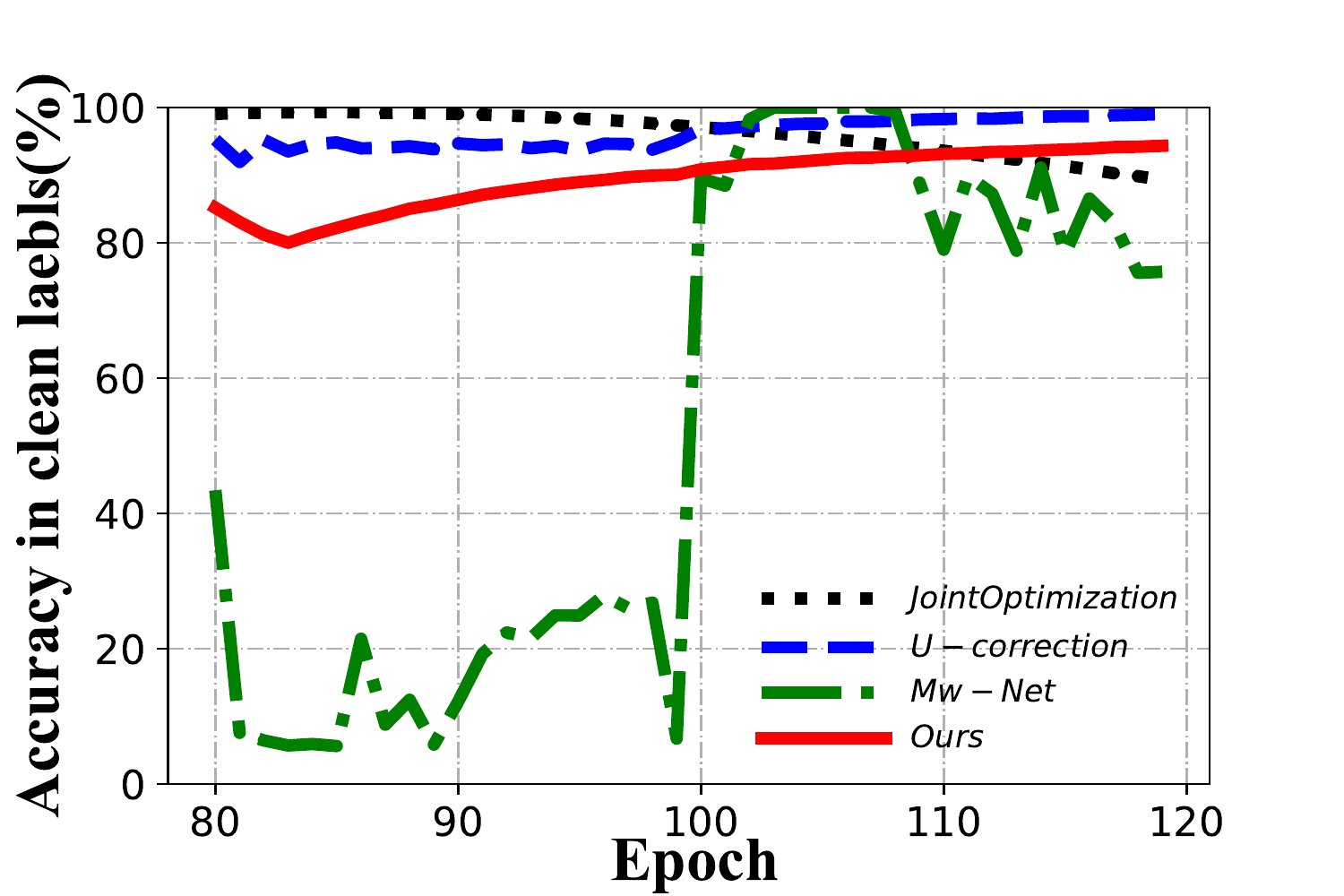}
  }
  \subfigure[Cifar100-Asy-Noise40$\%$]{
    %\label{fig:subfig:b} %% label for second subfigure
    \includegraphics[width=2.1in]{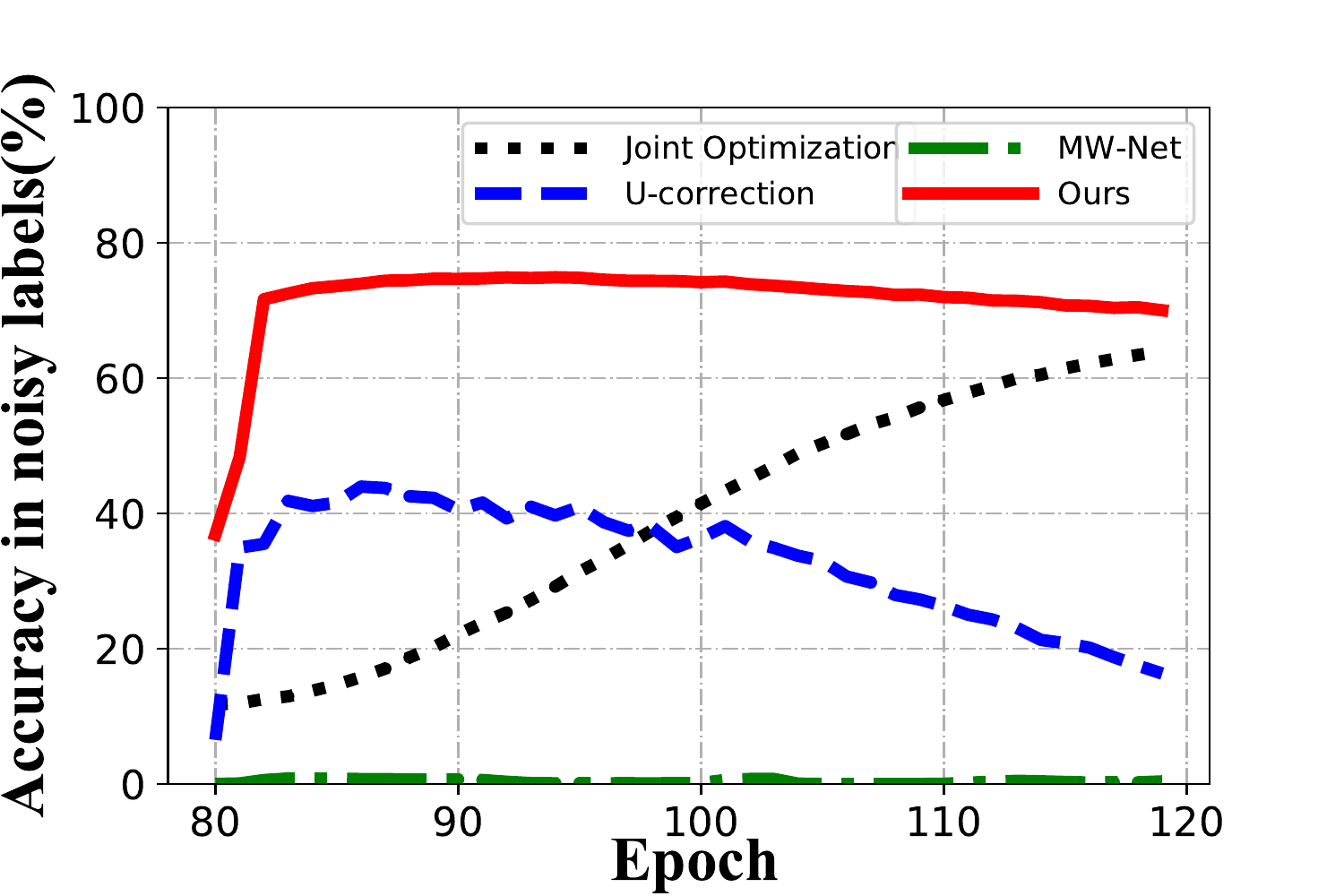}
  }
  \subfigure[Cifar100-Sym-Noise60$\%$]{
    %\label{fig:subfig:a} %% label for first subfigure
    \includegraphics[width=2.1in]{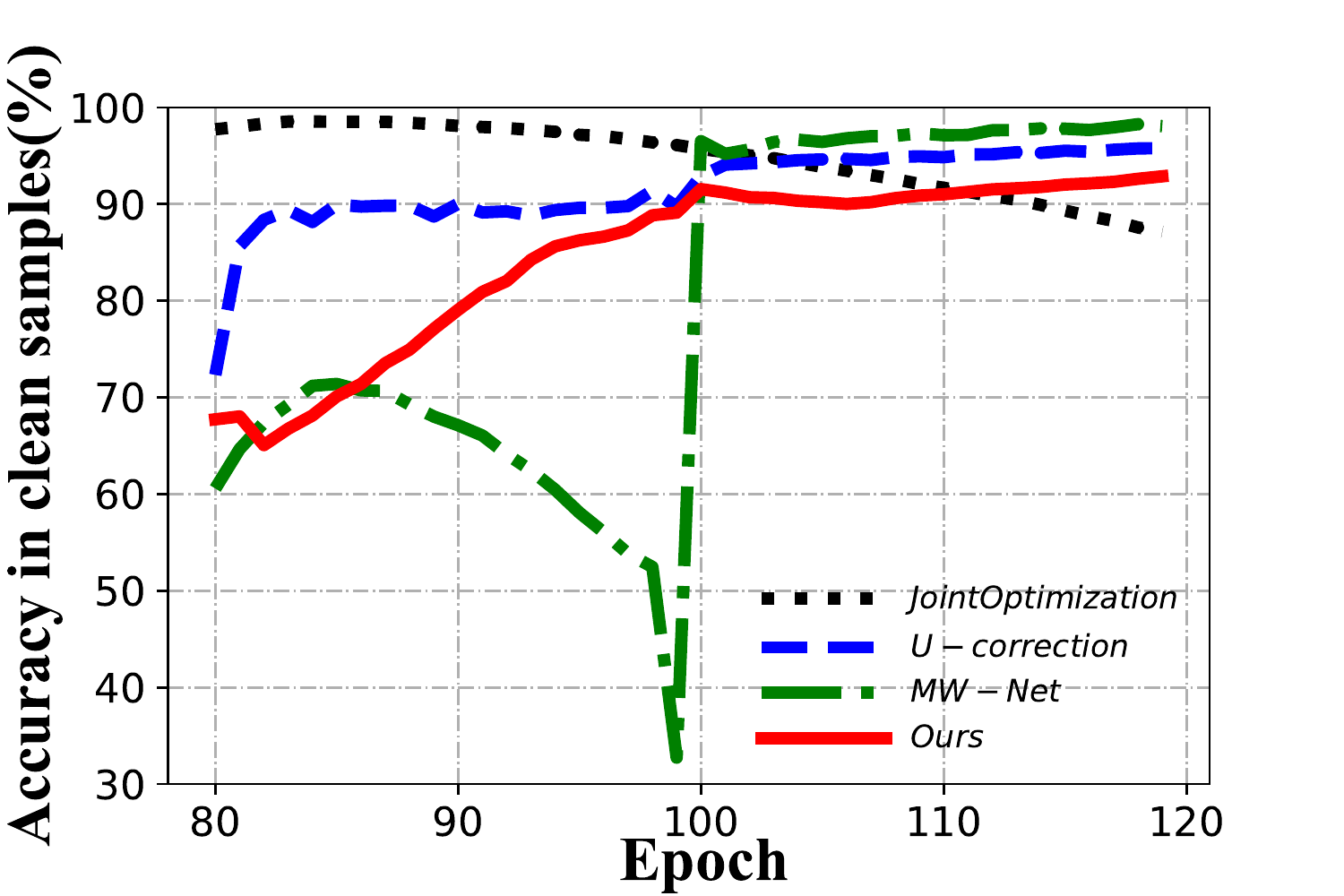}
  }
  \subfigure[Cifar100-Sym-Noise60$\%$]{
    %\label{fig:subfig:b} %% label for second subfigure
    \includegraphics[width=2.1in]{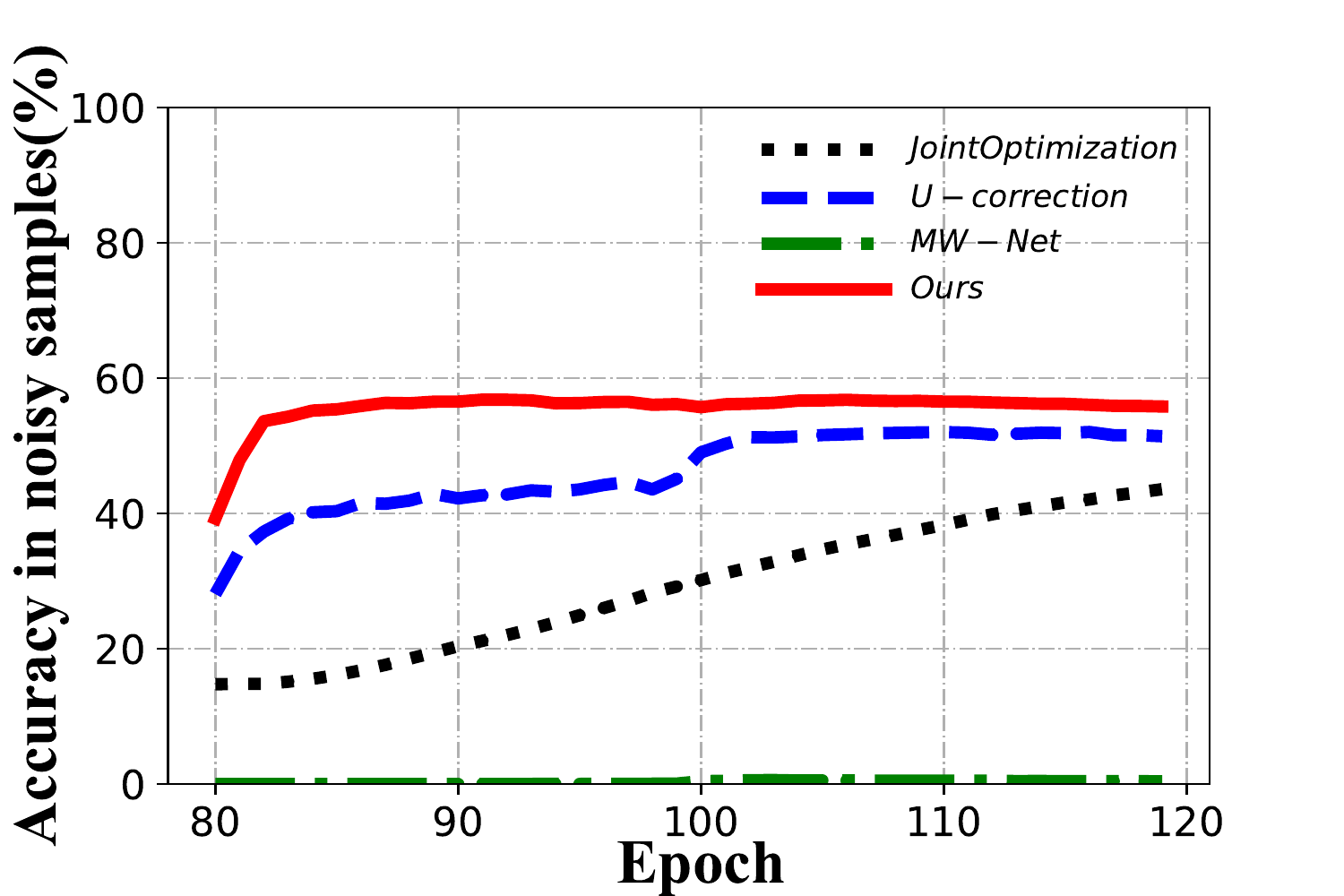}
  }
  \caption{The corrected label accuracy on different noise types and ratios. (a) and (b) show the accuracy in clean and noisy samples, respectively, under $40\%$ asymmetric noise on CIFAR100. (c) and (d) show the accuracy in clean and noisy samples, respectively, under $60\%$ symmetric noise on Cifar100.}
  \label{LA}
  %\label{fig:subfig} %% label for entire figure
\end{figure}
\section{More Setting Details on Our Method}
\textbf{Network Structure.} \quad For the classifier network, we choose ResNet-34. For meta learners, inspired by \cite{shu2019meta}, we adopt a single multilayer perceptron (MLP) with one hidden layer containing 100 nodes in both networks $\alpha(\cdot)$ and $\beta(\cdot)$ to output the weight.

\textbf{Synthetic Datasets.} \quad We conducted these experiments across both synthetic datasets (i.e. CIFAR-10 and CIFAR-100 with different types and levels of noise) sharing the same configuration and lead to consistent improvements over the state-of-the-arts. Our proposed meta label corrector was trained with two steps, firstly through warm-up to learn the structured data  with only cross-entropy loss, and secondly by introducing two meta learners to correct labels under the guidance of a small set of meta data with clean labels. We used SGD with a momentum of 0.9, a weight decay of $5\times10^{-4}$, and the batchsize of 100. The learning rate is set as 0.1 which is divided by 10 after 80 and 100 epochs for a total of 120 epochs. After we trained the first step with 80 epochs, we used $\eta_{1}=0.001,\eta_{2}=0.001$ and Adam to train the two meta learners.

\textbf{Clothing1M Data.} \quad
In training on the Clothing1M dataset, we used ResNet-50 pre-trained on ImageNet to align experimental condition with previous study \cite{patrini2017making,tanaka2018joint}. We resized the images as $256\times256$, performed mean subtraction and cropped the middle $224\times 224$ for preprocessing. For classifier network, we used SGD with a momentum of 0.9, a weight decay of $10^{-3}$, and batch size of 32. The initial learning rate is set as $8 \times 10^{-4}$ and divided by 10 after 5 epochs. We trained the network for 10 epochs and began updating labels from the 2nd epoch (i.e. we only warm up for the 1st epoch). For meta learners $\alpha(\cdot),\beta(\cdot)$, we set the initial learning rate $\eta_{1}=0.001,\eta_{2}=0.001$ and used Adam to optimize the training process.

\textbf{Accuracy of Corrected Label.} \quad
In section 3.1, we plot the accuracy of corrected labels to show the effectiveness of our proposed method. Since both JointOptimization \cite{tanaka2018joint} and U-correction \cite{arazo2019unsupervised} have warm-up operations, for more comprehensive comparison, we let the three methods (i.e. JointOptimization, U-correction, Ours) begin with correct labels from the 80th epoch. For MW-Net, we just follow its original settings (i.e. starts sample-reweighting from the 1st epoch without warm-up). We normalized its sample weights, and consider those samples which weight greater than 0.5 as its preserved clean samples (the rest samples are corrupted ones). From the perspective of label correction, its accuracy of the corrected label is the proportion of clean samples it retains on the original clean ones.
%inspired by \cite{shu2019meta} we adopt a single multilayer perceptron (MLP) with one hidden layer containing 100 nodes to output the weight for i -th training sample

\section{More Experimental Results}
To further analyze the corrected label accuracy of different methods as demonstrated in Section 3.1, we plot Fig.\ref{LA} to show the accuracy in clean/noisy labels concretely. Fig.\ref{LA}(b)(d) reflect how many original clean samples are rectified mistakenly, and Fig.\ref{LA}(a)(c) represent how many original noisy samples are corrected accurately.

It can be seen that the accuracy in noisy labels decreases in the training process of U-correction \cite{arazo2019unsupervised} from Fig.\ref{LA}(b). That is because it used unsupervised clustering method to split the clean and noisy samples, which is easy to treat most samples as clean ones when it processes imbalanced data that the clean samples are the majority. Joint Optimization \cite{tanaka2018joint} performs well on the correction of noisy samples, as shown in Fig.\ref{LA}(b)(d). By simply using predicted labels to replace the original ones, this strategy, however, also causes the critical issue that many original clean samples are corrupted, as shown in Fig.\ref{LA}(a)(c).

%the proportion of clean samples through corrected on the whole original noisy samples.

\end{appendix}

\end{document}